\documentclass{article}
\usepackage{iclr2027_conference,times}
\usepackage{natbib}
\usepackage{array}
\usepackage{latexml}
\iflatexml\providecommand{\arraybackslash}{}\fi
\providecommand{\iclrfinalcopy}{}
\usepackage{graphicx}
\iflatexml
\renewcommand{\resizebox}[3]{#3}
\renewcommand{\rotatebox}[2]{#2}
\renewcommand{\parbox}[3][]{\par\lxWithClass{ltx_align_justify}{#3}\par}
\fi
\usepackage{booktabs}
\usepackage{tabularx}
\usepackage{amsmath}
\usepackage[section]{placeins}
\usepackage[colorlinks=true,linkcolor=blue,citecolor=blue,urlcolor=blue]{hyperref}
\usepackage{url}
\usepackage{caption}

\newcommand{\fullmark}{\ensuremath{\bullet}}
\newcommand{\partialmark}{\ensuremath{\odot}}
\newcommand{\openmark}{\ensuremath{\circ}}

\title{EmoTrace-Bench: Diagnosing Emotional Expression in Text-to-Video Generation through Controlled Cues and Intensity}

\iflatexml
\author{Yeheng Wang\\{Communication University of China, Beijing, China}\\{Equal contribution}
\and Liudan Shi\\{Communication University of China, Beijing, China}\\{Equal contribution}
\and Yi Zheng\\{Communication University of China, Beijing, China}\\\texttt{zhengyi7592@cuc.edu.cn}\\{Corresponding author}
\and Jingke Li\\{Kling Team, Beijing, China}
\and Shiyuan Liu\\{Kling Team, Beijing, China}
\and Xin Tao\\{Kling Team, Beijing, China}
\and Yuanxing Zhang\\{Kling Team, Beijing, China}
\and Guoyu Sun\\{Communication University of China, Beijing, China}}
\else
\author{Yeheng Wang$^{1*}$\quad Liudan Shi$^{1*}$\quad Yi Zheng$^{1\dagger}$\quad Jingke Li$^{2}$\quad Shiyuan Liu$^{2}$\\
\textbf{Xin Tao$^{2}$\quad Yuanxing Zhang$^{2}$\quad Guoyu Sun$^{1}$}\\[4pt]
\normalfont $^{1}$Communication University of China, Beijing, China\\
\normalfont $^{2}$Kling Team, Beijing, China\\[2pt]
\normalfont\small $^{*}$Equal contribution.\quad $^{\dagger}$Corresponding author: \texttt{zhengyi7592@cuc.edu.cn}.}
\fi

\iclrfinalcopy
\begin{document}

\maketitle

\begin{abstract}

Text-to-video (T2V) models are increasingly used to generate videos with targeted emotional expressions, yet existing evaluation methods mainly rely on holistic video--text alignment and provide limited insight into whether failures arise from target-emotion expression, intensity control, or specific affective cues. We introduce EmoTrace-Bench, a fine-grained diagnostic benchmark that controls emotion intensity and selected affective cues in prompts and evaluates generated videos at multiple levels. EmoTrace-Bench contains 448 controlled prompts covering eight emotions, two intensity levels, and diverse cue conditions. Using 12 T2V models, we generate 5,376 videos and evaluate target-emotion presence, perceived intensity, and ten audiovisual dimensions of emotional expression with an automated evaluator calibrated and validated against human annotations on a subset of the data. Results reveal substantial differences across models and prompt designs. Prompts that explicitly describe a character's emotional reaction are followed more reliably than prompts that describe only events or scenes, whereas adding more affective cues or using more explicit cinematic audiovisual language does not consistently improve emotional generation. These results illustrate how aggregate alignment scores may overlook localized differences in target-emotion expression, intensity control, and cue realization. EmoTrace-Bench provides a traceable and interpretable framework for diagnosing emotional video generation without modifying existing generation pipelines, and we release the prompts and the automated evaluation rules.

\end{abstract}

\section{Introduction}

\begin{figure}[t]
    \centering
    \captionsetup{font=small}
    \includegraphics[width=\textwidth]{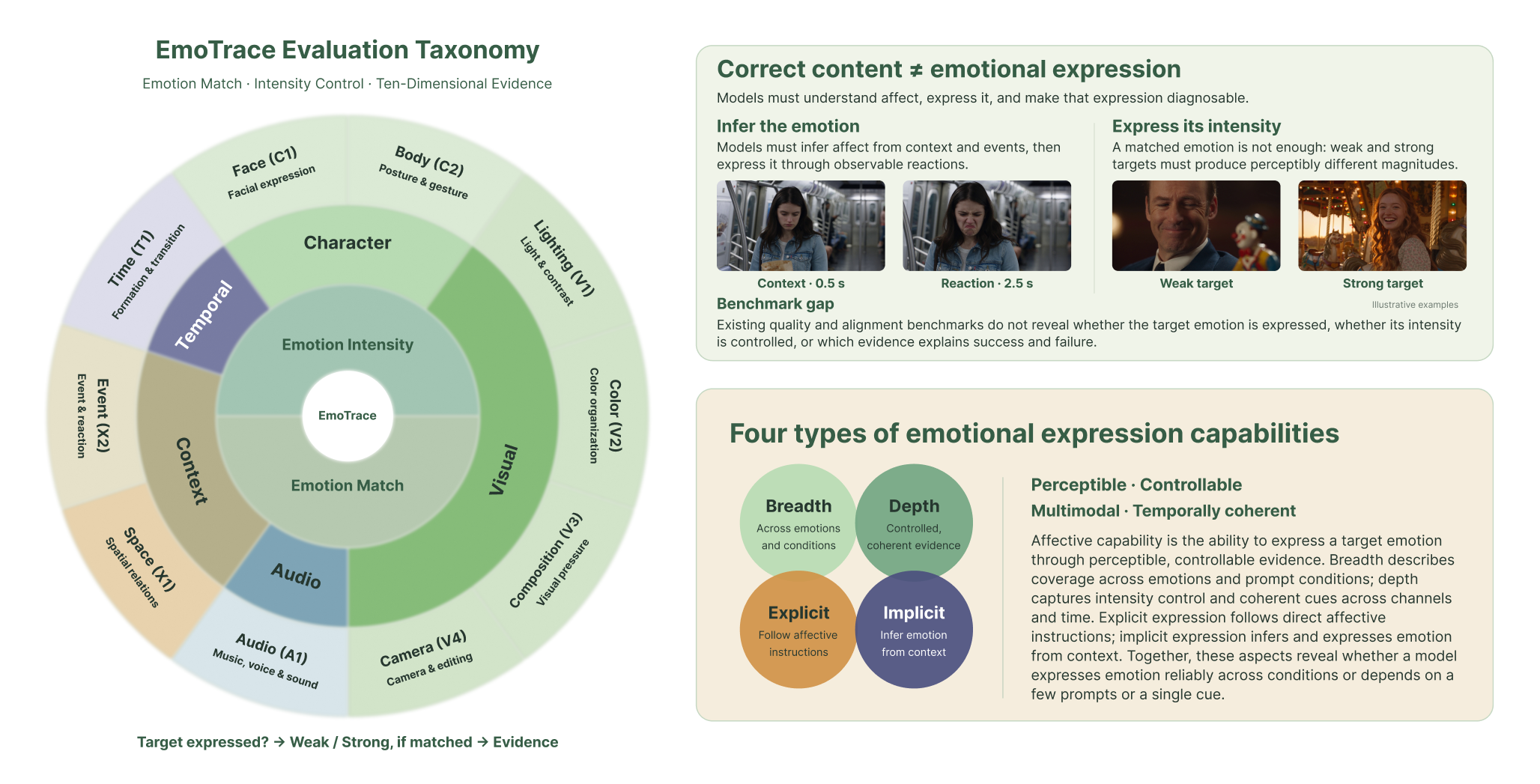}
    \caption{EmoTrace-Bench at a glance. Left: the three-level evaluation taxonomy---whether the target emotion is expressed, whether its perceived intensity is weak or strong, and which of ten expression channels (character, visual form, audio, context, temporal) provide the evidence. Top right: correct content does not guarantee emotional expression; a model must infer the emotion from context and reaction and realize the requested intensity, and existing benchmarks do not diagnose where this fails. Bottom right: the four capability facets examined in Section~4---breadth across emotions and prompt conditions, depth of intensity control and cue support, and explicit versus implicit cue expression.}
    \label{fig:teaser}
\end{figure}

\suppressfloats[t]

Generating videos with explicit emotional expression is an important capability for controllable content generation in text-to-video (T2V) models. For controllable emotion generation, the key question is not only whether a video presents the target emotion, but also whether it follows the prompt's requirements regarding the emotion category, emotion intensity, and specific expressive cues. Emotion generation in T2V is therefore not a single-dimensional problem; rather, it simultaneously involves multiple interrelated yet distinguishable capabilities, including emotion category, emotion intensity, and specific emotional cues.

However, existing T2V evaluations have difficulty effectively distinguishing these capabilities. Existing multidimensional video generation benchmarks mainly focus on semantic consistency, visual quality, motion performance, and compositional relationships, and they generally lack explicit modeling of emotional expression \citep{vbench,vbenchpp,vbench2,evalcrafter,videobench}. The few works oriented toward emotion generation also mostly use aggregate scores to measure the overall match between generated videos and emotion prompts \citep{emovid}. Although these metrics can indicate whether a video aligns with its prompt overall, they provide limited insight into the specific causes of mismatches and may collapse qualitatively different generation errors into similar aggregate scores, thereby obscuring meaningful differences in models' ability to follow fine-grained emotional instructions. As a result, such evaluations cannot fully characterize a model's controllability over emotion category, intensity, and mode of expression in practical content creation.

Adding more output-side metrics alone is insufficient, because confounds often arise during prompt construction. Emotional prompts typically combine multiple cues, making it difficult to attribute output changes to specific factors. Differences across emotion categories can further confound intensity comparisons. Reliable diagnosis therefore requires controlled inputs as well as fine-grained outputs.

\begin{figure}[t]
    \centering
    \captionsetup{font=small}
    \includegraphics[width=\textwidth]{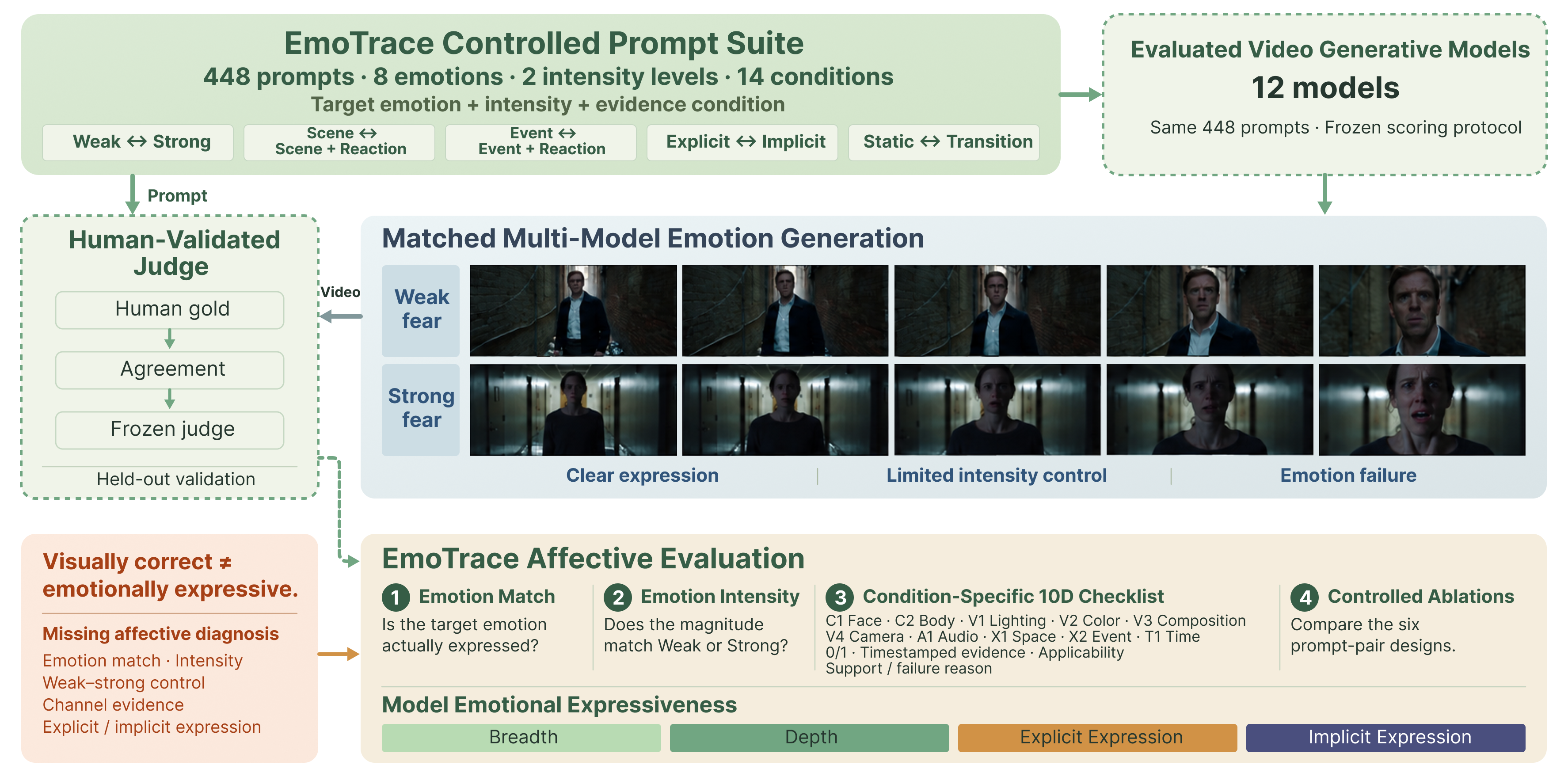}
    \caption{Overview of the EmoTrace-Bench pipeline. Prompts from the controlled suite (top) are sent unchanged to every model; the middle row shows matched weak/strong generations for one fear prompt pair across models. All videos are scored by a single automated evaluator that is frozen after calibration against human labels (left) and applied through the two-stage affective evaluation (bottom).}
    \label{fig:pipeline}
\end{figure}

We propose EmoTrace-Bench (Figure~\ref{fig:teaser}), which treats prompts as controlled experimental variables. It varies two main factors: affective cues and emotion intensity. Cue-controlled prompts change only one cue phrase while keeping the rest fixed, with additional paraphrased variants to test sensitivity to wording. Intensity-controlled prompt pairs differ only in intensity-related terms while preserving the emotion category, scene, and other content. This design enables separate analysis of emotion category, intensity, and specific cues. Figure~\ref{fig:pipeline} summarizes the overall pipeline.

At the output, we annotate target-emotion success, perceived intensity, and ten audiovisual cue dimensions. Using 448 controlled prompts, we generate 5,376 videos from 12 T2V models and collect human annotations for 1,120 videos from three of them to validate the automated evaluation. The analysis reveals that emotion-category generation and intensity control are distinct capabilities, model performance can be sensitive to cue wording, and automated evaluation agrees broadly with human judgments while small systematic differences remain on some fine-grained dimensions.

By controlling variables at the input and conducting fine-grained diagnosis at the output, EmoTrace-Bench decomposes emotion-generation performance that was previously mixed together in overall scores into more specific dimensions and prompt factors. At the same time, through per-video, per-attribute comparisons between human and automated evaluation, it further reveals the scope of applicability and limitations of automated evaluation across different emotional dimensions.

The main contributions of this paper are threefold:
\begin{itemize}
\item[(i)] We introduce \textbf{EmoTrace-Bench}, a fine-grained diagnostic benchmark for emotion generation in text-to-video models. It decomposes emotion generation into three complementary levels: emotion category, emotion intensity, and ten affective cue channels, enabling localized diagnosis of generation failures.

\item[(ii)] We develop a controlled prompt-construction framework that independently varies emotion intensity and affective cues under a minimal-difference design. It also includes semantically matched cue variants with different linguistic realizations, allowing us to study prompt sensitivity while reducing confounding across input factors.

\item[(iii)] We benchmark 12 text-to-video models on 448 controlled prompts, yielding 5,376 generated videos, and collect human annotations for 1,120 videos to validate the automated evaluation protocol. Our analysis reveals substantial differences across models in emotion-category generation, intensity control, and prompt sensitivity, while showing that aggregate automated judgments are broadly aligned with human annotations. We release the prompts and the automated evaluation rules.

\end{itemize}

\section{Related Work}

\subsection{Text-to-Video Generation Evaluation}

T2V evaluation has evolved from overall quality assessment toward multidimensional diagnosis. VBench, VBench++, VBench-2.0, EvalCrafter, and Video-Bench evaluate visual quality, motion, temporal consistency, subject consistency, spatial relations, text alignment, and higher-level generation fidelity \citep{vbench,vbenchpp,vbench2,evalcrafter,videobench}. VABench and T2AV-Compass further consider audio quality and audiovisual consistency \citep{vabench,t2avcompass}, while T2V-CompBench uses controlled prompts to evaluate compositional generation \citep{t2vcompbench}.

Despite this progress, these benchmarks mainly target general generation quality, semantic adherence, and compositionality. Emotional expression is rarely modeled as a structured capability with separable dimensions. Existing evaluations therefore provide limited diagnosis of whether errors arise from emotion category, intensity, or specific character, scene, and audiovisual cues.

\subsection{Emotional Video Generation and Controllable Expression}

Recent work has begun to study emotion-aware video generation. EmoVid provides a multimodal dataset for emotion-centric video understanding and generation \citep{emovid}; FG-ECVG explores fine-grained controllable emotional video generation \citep{fgecvg}; and EmoTalker studies editable emotional expression in talking-face generation \citep{emotalker}. EmoWorld further disentangles atmosphere, affective semantic cues, and temporal progression, showing that video emotion contains multiple controllable components \citep{emoworld}.

These works mainly focus on emotional data, generation, or model-side control. Their evaluations still commonly rely on emotion-recognition accuracy, condition matching, or aggregate scores. Such metrics may assign similar scores to different failures, such as generating the correct emotion at the wrong intensity or creating an appropriate atmosphere without the requested facial or bodily expression. Thus, an overall emotion score is insufficient for diagnosing fine-grained emotional controllability.

\subsection{Fine-Grained Emotional and Audiovisual Cues}

Emotion is conveyed through multiple channels. Psychological studies highlight the roles of facial expression, body posture, and movement \citep{ekman1978facs,cohn2007observer,coulson2004attributing,dael2012body,dael2012emotion,aviezer2012body}, while film studies show that shot scale, motion, lighting, color, sound, and music also shape emotional perception \citep{balint2018shot,detenber1998roll,chion1994audio,cohen2010music,bordwell2020film}. Recent benchmarks such as Stable Cinemetrics and FilmBench evaluate structured cinematic attributes \citep{stablecinemetrics,filmbench}, while HVEval and HUG-VIS examine human-centered appearance, action, and emotion \citep{hveval,hugvis}.

Building on these directions, EmoTrace-Bench decomposes emotion generation into emotion category, emotion intensity, and specific affective cues. It combines controlled prompts with fine-grained output evaluation to diagnose whether the target emotion is generated, whether its intensity follows the prompt, and through which character, scene, and audiovisual cues it is expressed. This moves emotion-generation evaluation from holistic matching toward traceable fine-grained diagnosis.

Table~\ref{tab:benchmark-comparison} summarizes these differences in affective targets, controlled inputs, diagnostic outputs, and human validation.

\begin{table*}[t]
\centering
\scriptsize
\setlength{\tabcolsep}{3.2pt}
\renewcommand{\arraystretch}{1.12}
\caption{\textbf{Capability coverage of representative video-generation
benchmarks.} Filled, dotted, and open circles denote full, partial or
domain-restricted, and absent support, respectively.}
\label{tab:benchmark-comparison}
\resizebox{\textwidth}{!}{%
\begin{tabular}{@{}l*{12}{c}@{}}
\toprule
& \multicolumn{3}{c}{\textit{Affective target}}
& \multicolumn{3}{c}{\textit{Controlled input}}
& \multicolumn{3}{c}{\textit{Diagnostic output}}
& \multicolumn{3}{c}{\textit{Human validation}} \\
\cmidrule(lr){2-4}\cmidrule(lr){5-7}\cmidrule(lr){8-10}\cmidrule(lr){11-13}
Benchmark
& EC & EI & AC
& CI & IP & GF
& CL & EL & DS
& PL & SV & MR \\
\midrule
EmoVid~\citep{emovid}
& \fullmark & \openmark & \partialmark
& \openmark & \openmark & \partialmark
& \openmark & \fullmark & \fullmark
& \openmark & \partialmark & \fullmark \\
HVEval~\citep{hveval}
& \partialmark & \openmark & \partialmark
& \openmark & \openmark & \partialmark
& \openmark & \partialmark & \fullmark
& \openmark & \partialmark & \fullmark \\
HUG-VIS~\citep{hugvis}
& \fullmark & \openmark & \openmark
& \openmark & \openmark & \partialmark
& \openmark & \fullmark & \fullmark
& \openmark & \partialmark & \fullmark \\
\addlinespace[2pt]
Stable Cinemetrics~\citep{stablecinemetrics}
& \openmark & \openmark & \partialmark
& \partialmark & \openmark & \fullmark
& \partialmark & \openmark & \fullmark
& \openmark & \partialmark & \partialmark \\
FilmBench~\citep{filmbench}
& \partialmark & \openmark & \partialmark
& \openmark & \openmark & \partialmark
& \openmark & \partialmark & \fullmark
& \openmark & \partialmark & \fullmark \\
T2V-CompBench~\citep{t2vcompbench}
& \openmark & \openmark & \openmark
& \openmark & \openmark & \fullmark
& \openmark & \openmark & \fullmark
& \openmark & \fullmark & \fullmark \\
\addlinespace[2pt]
VBench family~\citep{vbench,vbenchpp,vbench2}
& \openmark & \openmark & \openmark
& \openmark & \openmark & \fullmark
& \openmark & \openmark & \fullmark
& \openmark & \openmark & \fullmark \\
Video-Bench~\citep{videobench}
& \openmark & \openmark & \openmark
& \openmark & \openmark & \openmark
& \openmark & \openmark & \fullmark
& \openmark & \fullmark & \fullmark \\
VABench~\citep{vabench}
& \openmark & \openmark & \openmark
& \openmark & \openmark & \partialmark
& \openmark & \openmark & \fullmark
& \openmark & \partialmark & \fullmark \\
T2AV-Compass~\citep{t2avcompass}
& \openmark & \openmark & \openmark
& \openmark & \openmark & \partialmark
& \openmark & \openmark & \fullmark
& \openmark & \partialmark & \fullmark \\
EvalCrafter~\citep{evalcrafter}
& \openmark & \openmark & \openmark
& \openmark & \openmark & \partialmark
& \openmark & \openmark & \fullmark
& \openmark & \openmark & \fullmark \\
\midrule
\textbf{EmoTrace-Bench (ours)}
& \fullmark & \fullmark & \fullmark
& \fullmark & \fullmark & \fullmark
& \fullmark & \fullmark & \fullmark
& \fullmark & \fullmark & \partialmark \\
\bottomrule
\end{tabular}
}
\vspace{2pt}

\parbox{\textwidth}{\scriptsize
\textbf{Affective target:} EC = emotion category, EI = perceived emotion
intensity, AC = fixed affective-cue taxonomy.
\textbf{Controlled input:} CI = affective-cue isolation, IP = minimally
differing intensity pairs, GF = general factor control.
\textbf{Diagnostic output:} CL = per-video cue-channel labels, EL = per-video
emotion labels, DS = dimension- or criterion-level scores.
\textbf{Human validation:} PL = per-attribute label agreement, SV = sample- or
video-level validation, MR = model-ranking validation. Prompt-side control is
required for full credit under CI and IP; capture-side factorial designs receive
partial credit.}
\end{table*}

\section{Method}
\label{sec:method}

\subsection{Evaluation Target Design}
\label{subsec:evaluation-target-design}

EmoTrace-Bench evaluates emotional video generation at three levels: target emotion, perceived intensity, and affective cues. We first define the target emotion set and then specify the audiovisual dimensions used for fine-grained diagnosis.

\textbf{Selection of Eight Emotions.}
EmoTrace-Bench includes eight emotions: amusement, excitement, awe, contentment, anger, fear, disgust, and sadness. The selection follows the discrete emotion set established by Mikels et al.\ using the International Affective Picture System (IAPS) \citep{mikels2005emotional}. It contains four positive and four negative emotions spanning different typical arousal levels \citep{russell1980}.

\textbf{Three-Level Evaluation.}
For each generated video, the first level determines whether the target emotion is expressed. The second evaluates perceived emotion intensity as strong or weak when the target emotion is present, and as ``not matched'' otherwise. The third identifies which audiovisual cue dimensions support the target emotion. These levels capture whether the intended emotion is expressed, how strongly it is expressed, and through which cues.

\textbf{Multidimensional Affective Cues.}
Emotional expression in video can arise from character behavior, environmental context, and audiovisual form. Drawing on theories of emotional expression, film form, and an external inventory of affective cues \citep{smith2003film}, we define ten dimensions: facial expression, bodily behavior, lighting, color, composition, visual dynamics, temporal organization, sound, scene, and event. We retain cues that can be directly observed and judged relatively independently in short videos, while excluding factors that require long-form narrative, dialogue semantics, or broader temporal context. Operational definitions and channel-specific prerequisites are provided in Appendix~\ref{app:scoring} (Table~\ref{tab:appendix-cue-definitions}).

\subsection{Controlled Prompt Set}
\label{subsec:controlled-prompt-set}

EmoTrace-Bench uses controlled prompts to vary emotion intensity, affective cues, and prompt formulation while minimizing unrelated changes across conditions.

\textbf{Scale and Structure.}
Each prompt is represented as
\[
P(e,s,i,t)
=
B(e,s)
+
I(e,i)
+
\sum_{d \in \mathcal{M}_t} C_{d,t}(e,s),
\]
where \(e\) is the target emotion, \(s\) the scene, \(i\) the intensity level, and \(t\) the prompt template. \(B(e,s)\) is the shared base description, \(I(e,i)\) specifies emotion intensity, and \(C_{d,t}(e,s)\) denotes the realization of affective cue \(d\) under template \(t\). The template determines which cue segments are included through \(\mathcal{M}_t\) and, for selected conditions, how the same affective content is linguistically expressed.

The benchmark crosses 8 emotions, 2 intensity levels with 2 samples each, and 14 prompt templates, yielding 448 prompts. The templates include emotion-only prompts, individual facial, bodily, scene, and event cues, combinations of character and environmental cues, alternative linguistic realizations, facial changes, empty scenes, and prompts containing all explicitly writable cues. Eight of the ten evaluation dimensions---facial expression, bodily behavior, lighting, color, composition, visual dynamics, scene, and event---can be explicitly specified in prompts. Temporal organization and sound are evaluated only in the generated video. Each prompt records whether a cue dimension is explicitly specified or left unspecified but observable, allowing explicit instruction following to be distinguished from cues that emerge without being requested. Representative prompt conditions are shown in Figure~\ref{fig:prompt-design} in Appendix~\ref{app:prompt-construction}.

\textbf{Alternative Linguistic Realizations.}
To examine sensitivity to prompt formulation, we construct controlled prompt pairs that express similar affective content using different linguistic realizations. These pairs preserve the target emotion, scene, and intended affective content while changing how that content is described. They therefore allow us to test whether semantically similar instructions lead to different generation outcomes.

\textbf{Prompt Control and Validation.}
We apply three constraints to reduce unintended variation. First, strong--weak pairs share the same emotion, scene, base description, and template and differ only in preregistered intensity expressions; all \(224/224\) pairs pass character-level structural verification. Second, cue segments are reused across templates whenever possible. Of 384 reused segments, 372 are identical and 12 differ only in pronoun disambiguation; all 384 match after normalization. Third, controlled template pairs vary only the preregistered prompt factor, and all 448 prompts have unique SHA-256 hashes. These checks (Table~\ref{tab:prompt-audit}) establish structural consistency for controlled comparisons.

\subsection{Automated Evaluation and Human Annotation Protocol}
\label{subsec:evaluation-protocol}

Automated evaluation and human annotation follow the same three-level decision structure and information boundaries.

\textbf{Shared Evaluation Protocol.}
Both evaluators first determine whether the target emotion is present. If present, perceived intensity is labeled as strong or weak; otherwise it is labeled ``not matched.'' Each of the ten affective cue dimensions is then judged independently as present or absent in support of the target emotion. Emotion and intensity are judged directly and are not inferred from cue labels.

Both evaluators see only the generated video, including its native audio, and the name and definition of the target emotion. They do not have access to the generation prompt or other metadata. The target emotion is provided because the task evaluates whether a specified emotion is expressed. The target intensity is hidden to reduce label-following bias.

\textbf{Automated Evaluation.}
All 5,376 videos are evaluated once using Gemini~3.1 Pro with temperature 0 under a frozen two-stage protocol: the first stage judges target emotion and perceived intensity, and the second stage judges the ten cue channels (Appendix~\ref{app:scoring}). Videos are sampled at 8~fps. Auxiliary programs provide measurement summaries for visual dynamics and sound using cross-frame motion and audio features, respectively. The resulting single-run outputs are used for all 12-model benchmark results.

To assess evaluator stability, we conduct two additional evaluation runs on the 896-video development set. Development-set validity results are reported as the mean over the three runs (Appendix~\ref{app:validity}). Wan-2.2, HunyuanVideo-1.5, Kandinsky-5.0-T2V-Pro, and LongCat-Video-13.6B do not provide usable native audio, so the sound dimension and Full 10D aggregate are not reported for these models.

\textbf{Human Reference and Consensus.}
The human reference contains a development set for protocol calibration and a held-out test set for cross-generator validation. All first-pass annotations are collected without access to automated evaluation outputs.

The development set contains 896 videos from two generation models. Two senior annotators independently label all videos. Their matching labels form the first-pass consensus, while disagreements are resolved by the majority of available labels from five additional trained annotators (Appendix~\ref{app:validity}). The test set contains 224 videos from a third generation model and was not used in protocol development. Two trained annotators provide full first-pass annotations and form the initial consensus.

After the automated evaluator is frozen, senior annotators conduct a second review of both sets (Appendix~\ref{app:validity}). Reviewers see only the video and decision rules, without access to the generation prompt, model identity, target intensity, or automated outputs. Reviewed labels form the reviewed human reference, and fields without a reviewed label are excluded from analyses against it.

\textbf{Metric Computation.}
The joint emotion-and-intensity score is computed over all 448 videos per model. Visual 9D averages cue support over the nine non-audio channels and is available for all models; Full 10D adds the sound channel and is therefore reported only for audio-capable models (Appendix~\ref{app:scoring}). Both are computed only for videos that successfully express the target emotion and should therefore be interpreted together with target-emotion accuracy. ``Undecided'' and ``not applicable'' labels are excluded from the corresponding denominators.
\section{Results and Analysis}

We analyze EmoTrace-Bench from three perspectives. First, we compare 12 T2V models in terms of target-emotion generation, intensity control, and cue realization. Second, we use controlled prompt pairs to examine the effects of cue specification, cue combination, and linguistic realization. Third, we assess the reliability of the automated evaluator through comparisons with human judgments and an alternative evaluator backbone.

\subsection{Emotion Generation and Intensity Control}

The 12 models exhibit distinct capability profiles. Target-emotion accuracy ranges from 49.3\% to 87.0\%, joint emotion-and-intensity accuracy from 29.9\% to 55.1\%, and conditional Visual 9D expression depth from 62.7\% to 83.5\%. Paired intensity-control success is substantially lower, ranging from 10.3\% to 29.5\%. Taken together, these results suggest that models are generally more successful at expressing the target emotion than at reliably controlling its perceived intensity.

\begin{figure}[!htbp]
    \centering
    \captionsetup{font=small}
    \includegraphics[width=\textwidth]{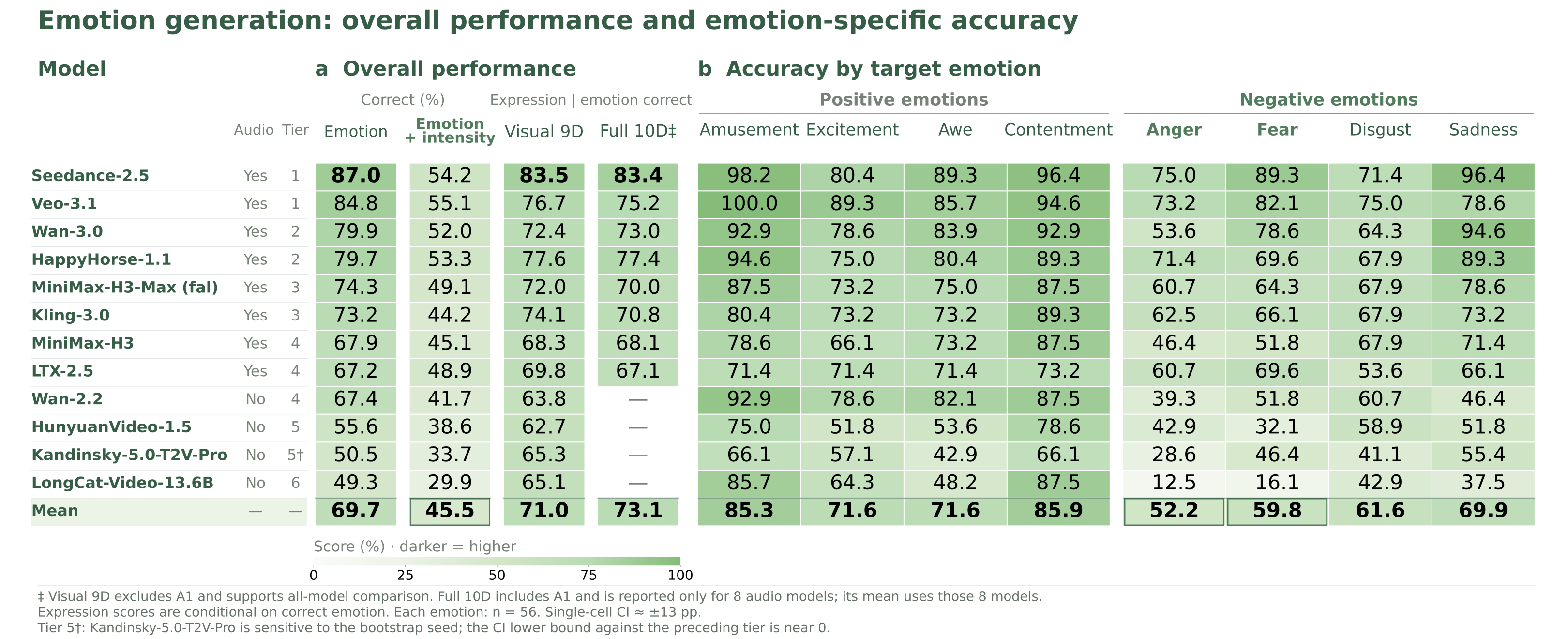}
    \caption{Emotion generation across 12 T2V models. (a) Overall performance: target-emotion accuracy (Emotion), joint emotion-and-intensity accuracy (Emotion + intensity; E+I in the text), Visual 9D and Full 10D cue support, and bootstrap group; models within a group are not ordered. (b) Target-emotion accuracy for each of the eight emotions. Scores are obtained from one blinded automated evaluation per video.}
    \label{fig:model-overall}
\end{figure}

Figure~\ref{fig:model-overall}(a) summarizes the overall model results. Prompt-level paired bootstrap analysis of target-emotion accuracy (Appendix~\ref{app:scoring}) yields six bootstrap-based groups of models for which no significant within-group differences are detected under the adopted procedure. Seedance-2.5 and Veo-3.1 form the highest-performing group.

These differences are not an artifact of generation seed. Regenerating all 448 videos for Kling-3.0 under three generation seeds, with prompts and all other exposed settings fixed, moves target-emotion accuracy by 3.6 points (72.5--76.1\%) and joint emotion-and-intensity accuracy by 3.8 points (45.5--49.3\%), with no pairwise difference significant under exact McNemar tests (Appendix~\ref{app:seed-sensitivity}). For this model and these three seed settings, the observed score ranges are approximately 7--11 times smaller than the corresponding cross-model spreads.

Performance also varies across emotion categories (Figure~\ref{fig:model-overall}(b)). Contentment (85.9\%) and amusement (85.3\%) achieve the highest average target-emotion accuracy, whereas anger (52.2\%) and fear (59.8\%) are the lowest. Across the emotion categories included in our benchmark, the four positive emotions achieve 17.7 percentage points higher target-emotion accuracy on average than the four negative emotions, with the difference consistent in direction across all 12 models.

Intensity control exhibits a different failure pattern. Across models, 17.9\%--48.2\% of strong--weak pairs receive the same perceived intensity, whereas reversals occur in at most 5.8\% of pairs (Table~\ref{tab:intensity-control}, Appendix~\ref{app:additional-results}). Thus, when intensity control fails, it is much more often because strong and weak prompts produce the same perceived intensity.

Cue-level results reveal additional cross-model differences (Figure~\ref{fig:implicit-cue-profiles}, left; Table~\ref{tab:ten-model-cues}, Appendix~\ref{app:additional-results}). Visual dynamics shows the largest variation across models, followed by event/reaction, sound, and lighting, whereas temporal organization and facial expression vary less. Because human--automated agreement is only moderate for composition and scene/space (Appendix~\ref{app:validity}), we report these two channels descriptively without further cross-model interpretation. 

Overall, target-emotion generation, intensity control, and cue realization capture different aspects of model behavior. These results motivate reporting these dimensions separately rather than collapsing them into a single aggregate score.

\subsection{Effects of Controlled Prompt Design}

We next examine how model outputs change under controlled prompt variations. The analysis focuses on three factors: explicit cue specification, cue combination, and linguistic realization. Per-model results for all 14 templates are reported in Table~\ref{tab:template-ei} (Appendix~\ref{app:additional-results}).

\paragraph{Explicit Cue Specification.}
Explicitly specifying character reactions produces the largest and most consistent gains. Adding a character reaction to a scene-only prompt improves joint emotion-and-intensity accuracy by 25--59 percentage points across models, with a mean gain of 45 points. Adding a reaction to an event prompt yields a 22--69 point gain, with a mean of 46 points. After Holm correction, these improvements are significant for 9/12 and 7/12 models, respectively. Current models therefore follow explicit facial and bodily reactions more reliably than they infer such reactions from scene or event context alone.

\begin{figure}[!htbp]
    \centering
    \captionsetup{font=small}
    \includegraphics[width=\textwidth]{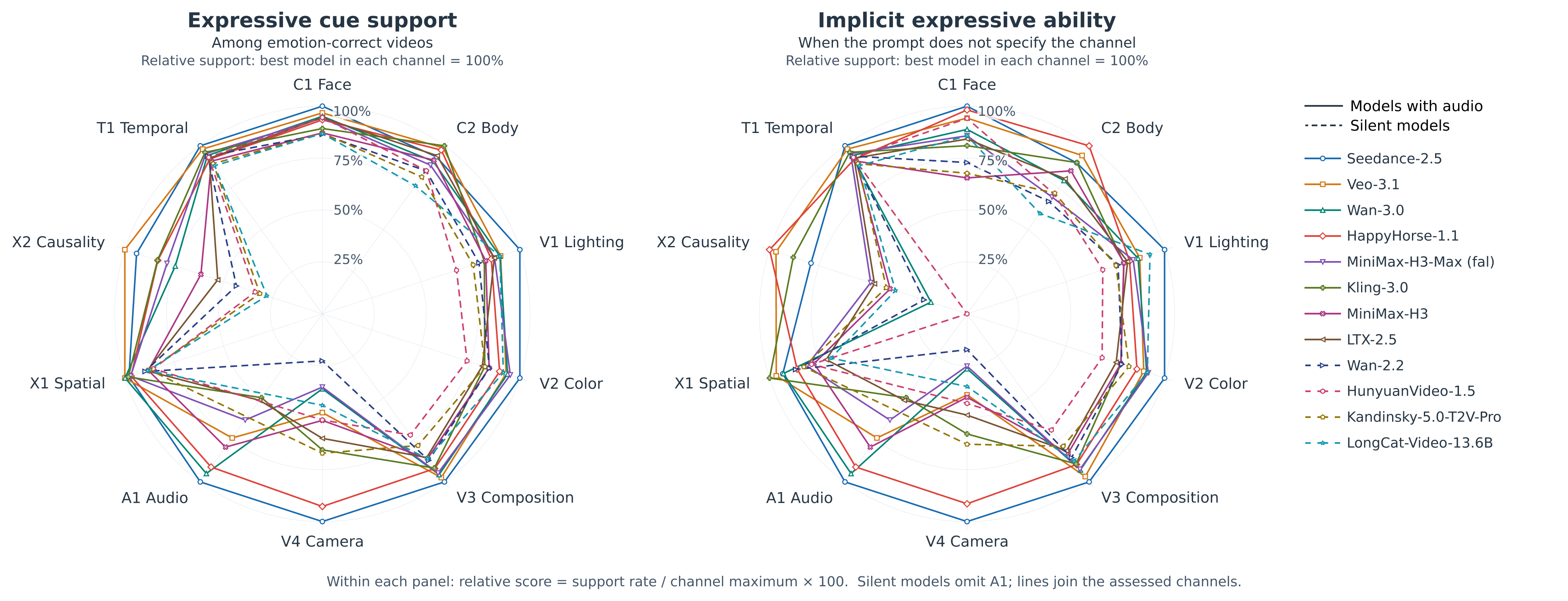}
    \caption{Overall (left) and implicit (right) cue-expression profiles of 12 T2V models. Values are relative to the best model in each channel.}
    \label{fig:implicit-cue-profiles}
\end{figure}

For cues that are not explicitly specified, we still record whether they appear in the generated video. Figure~\ref{fig:implicit-cue-profiles} compares overall cue support with support under unspecified conditions. These results characterize cue co-occurrence only and do not support causal interpretation.

\paragraph{Cue Combination.}
The benefit of adding a cue depends on which cue is added, not on how many are present. Adding a facial expression to a body-only prompt raises joint emotion-and-intensity accuracy from 40\% to 64\% on average (+24 points; significant for 2/12 models), whereas adding bodily behavior to a facial-expression prompt adds only 7 points (57\% to 64\%; significant for no model). Descriptively, facial expression contributes more than bodily behavior in this contrast. Likewise, the template that combines all explicitly writable cues (ALL\_DYNAMIC, 64\%) performs no better than the face-and-body prompt and only 2 points above the emotion-label baseline (EMO, 62\%; Table~\ref{tab:template-ei}).

\paragraph{Alternative Linguistic Realizations.}
Describing a scene by how it feels conveys emotion more effectively than specifying how it should be filmed. With a character present, experiential atmosphere descriptions reach 27\% joint emotion-and-intensity accuracy on average, compared with 13\% for matched cinematic instructions; 9/12 models favor the experiential form, 2/12 significantly after correction. In empty scenes the two formulations perform alike (38\% vs.\ 37\%). For short, single-scene prompts of this kind, experiential language elicited emotional expression more often in our data.

All pre-specified contrasts use exact paired McNemar tests with model-wise Holm correction (Table~\ref{tab:controlled-template-pairs}). Because each condition contains 32 videos (16 strong--weak pairs), we read these comparisons as diagnostic evidence of how models respond to specific prompt changes.

\subsection{Agreement Between Automated and Human Evaluation}

Because the preceding analyses rely on automated scores, we compare the automated evaluator with human judgments at both the annotation and model levels. The evaluation uses a development set of 896 videos and a test set of 224 videos from a third generation model, held out from protocol development.

\paragraph{Human Agreement.}
On the development set, pairwise human agreement is 85.7\% for target emotion, 77.9\% for intensity, and 80.8\% for the ten cue dimensions, with Fleiss' \(\kappa\) values of 0.65, 0.50, and 0.60, respectively \citep{fleiss1971}. On the test set, agreement between the two full-coverage annotators is 81.7\%, 75.5\%, and 80.2\%, with Cohen's \(\kappa\) values of 0.54, 0.50, and 0.58, respectively \citep{cohen1960}. These agreement levels characterize the reliability of the human annotations used to construct the reference labels for evaluating the automated judgments (Table~\ref{tab:human-reference} in Appendix~\ref{app:validity} and Table~\ref{tab:judge-validity}).

\begin{table}[!htbp]
\centering
\scriptsize
\setlength{\tabcolsep}{4.5pt}
\caption{Agreement between the automated evaluator and the human reference.}
\label{tab:judge-validity}
\begin{tabular*}{\textwidth}{@{\extracolsep{\fill}}lcccc@{}}
\toprule
Reference & Target emotion & Conditional intensity & Joint three-way & Ten dimensions \\
\midrule
Development, first-pass human reference & 88.1 / .67 & 79.1 / .58 & 73.4 / .59 & 83.0 / .63 \\
Test, first-pass human reference & 84.7 / .59 & 88.3 / .76 & 77.0 / .65 & 80.5 / .59 \\
Development, reviewed human reference & 89.9 / .72 & 93.0 / .86 & 84.8 / .76 & 85.1 / .67 \\
Test, reviewed human reference & 90.2 / .76 & 92.0 / .84 & 87.1 / .81 & 83.0 / .64 \\
\bottomrule
\end{tabular*}

\vspace{2pt}

\parbox{\textwidth}{\scriptsize
Values are agreement (\%) / Cohen's \(\kappa\). Conditional intensity is computed only when both sides judge the target emotion present. Joint three-way evaluation includes ``not matched,'' ``weak,'' and ``strong.'' Development-set values are means over three independent runs; full standard deviations and valid counts are reported in Table~\ref{tab:validity-full}. The first-pass reference is the annotator consensus before senior review; the reviewed reference is the final label set after a second review by senior annotators. Neither was produced with access to automated outputs.
}
\end{table}

\paragraph{Automated--Human Agreement.}
Table~\ref{tab:judge-validity} reports agreement against the first-pass and the reviewed human reference. Against the first-pass reference, agreement is substantial on target emotion and the ten cue dimensions (\(\kappa=0.59\)--0.67) and moderate to substantial on conditional intensity (\(\kappa=0.58\)--0.76); against the reviewed reference it is 2--14 points higher, reflecting the 308 and 297 fields revised at review and, on the test set, the 41 videos that receive a label only after review (Appendix~\ref{app:validity}).

\paragraph{Model-Level Consistency.}
Aggregate automated and human scores differ by at most 4 percentage points on the two development-set models and by 4--6 points on the test-set model, where automated scores are conservatively lower. Across model--emotion cells, Spearman correlation is 0.94 for target emotion and 0.90 for cue pass rates across model--emotion--dimension cells, indicating strong cell-level consistency across the three human-annotated generators and eight emotions even where absolute scores shift slightly.

\paragraph{Judge-Backbone Sensitivity.}
Applying an independent Qwen3.8-Omni-Flash evaluator to the 224 test-set videos under the same frozen protocol yields moderate to substantial agreement with Gemini across target emotion, joint intensity, the ten cue dimensions, and sound (\(\kappa=0.62\)--0.68), and comparable agreement with the first-pass human reference (Appendix~\ref{app:judge-backbone}). On this test-set generator, the evaluation patterns are therefore not tied to a single evaluator backbone; cross-evaluator replication of the full 12-model benchmark remains future work.

\section{Conclusion}

We introduce EmoTrace-Bench, a fine-grained diagnostic benchmark for emotion generation in text-to-video models. It evaluates target emotion, perceived intensity, and ten audiovisual cue dimensions using controlled prompts that vary intensity, cue specification, and linguistic realization. Experiments on twelve T2V models show that emotion generation, intensity control, and cue realization are distinct capabilities. Overall, EmoTrace-Bench provides a structured way to localize emotional generation failures to emotion category, intensity, and specific expressive cues, offering a more interpretable basis for evaluating and improving controllable emotional video generation.

\subsection*{AI use statement}

Generative AI systems played two distinct roles in this work. First, the 12
text-to-video models evaluated in the benchmark were the objects of study;
Gemini~3.1 Pro was used as the automated evaluator under the protocol described
in Section~\ref{subsec:evaluation-protocol}, and Qwen3.8-Omni-Flash (API ID:
\texttt{qwen3.8-omni-flash}) was used as a second evaluator in the backbone-sensitivity analysis
(Appendix~\ref{app:judge-backbone}). Their outputs were compared against human
annotations and were not treated as ground truth.

Second, generative AI tools assisted with language polishing of the authors'
finalized manuscript (improving clarity, grammar, and readability) and with
writing deterministic research scripts. The authors reviewed every
AI-assisted passage to confirm that it preserved the intended meaning of the
original text and manually reviewed all AI-assisted code. Generative AI was
not used to create the benchmark prompts, formulate the conceptual framework
or hypotheses, design the experiments or evaluation protocol, generate or
impute human annotations, or interpret the experimental results. The authors
take responsibility for the final content of the paper and all AI-assisted
artifacts.

\subsection*{Ethics statement}

The study evaluates model-generated videos and includes human annotation of
1,120 videos, of which 896 are used for development-set validity analysis and
224 form the test set. The annotation activity underwent a compliance review
by the Kling Team and was determined not to require formal review by an institutional
review board or independent ethics committee; this determination should not be
interpreted as a formal IRB exemption. All annotators were adults who were
informed about the task and participated voluntarily. They were compensated,
and they could skip content or withdraw from the task. None of the annotators,
including the two senior reviewers, are authors of this paper. The submission does not
include annotator-level provenance or annotator demographic information.
Because the benchmark covers anger, fear, disgust, and sadness, annotators could
encounter negatively valenced generated content.

EmoTrace-Bench may have dual-use implications: a diagnostic method for
identifying the cues that convey emotion could also be used to make generated
media more emotionally persuasive. We therefore recommend that cue-level
automated scores not be used as optimization targets without human validation.
The benchmark is intended for transparent evaluation and analysis of model
behavior rather than for profiling individuals or inferring emotions from real
people.

\subsection*{Reproducibility statement}

Section~\ref{sec:method} specifies the prompt construction, structural
verification, evaluation information boundaries, automated evaluation procedure,
human annotation protocol, aggregation rules, and metric denominators used in
this study. The project repository
(\url{https://github.com/Tonyheng6/EmoTrace-Bench}) contains all 448 prompts
and their per-prompt metadata, together with the frozen system prompts and
decision rules of the automated evaluator. Sample-level human reference labels
and annotator-level provenance are not released; the appendix reports the
consensus-construction procedure and aggregate agreement statistics. These
materials are provided so that the reported prompt structure and aggregate
results can be independently inspected.

\subsection*{Acknowledgments}

We thank all annotators for their careful work on the human reference labels.
We also thank the Kling Team for its support of this work.

\bibliography{iclr2027_conference}
\bibliographystyle{iclr2027_conference}

\clearpage
\appendix
\numberwithin{table}{section}
\numberwithin{figure}{section}
\section*{Appendix}

\section{Experimental Setup}
\label{app:experimental-setup}

\subsection{Models and Benchmark Scale}

We evaluate 12 publicly available text-to-video systems
(Table~\ref{tab:appendix-models}). Six open-weight systems were run locally
from their public releases; the remaining six were accessed through hosted
interfaces. MiniMax-H3 and MiniMax-H3-Max (fal) therefore differ in both the
checkpoint (H3-Max is a third-party post-trained variant of H3 served by fal)
and the inference channel, and the gap between them is not interpreted as a
pure model difference. We identify hosted systems by the public product name
available at the time of collection and open-weight systems by their public
release name. Exact provider-side checkpoint hashes and API revision strings
were not exposed consistently and are therefore not used to imply a finer
version match than the public names support.

The benchmark was frozen in September 2026. The current release records the
benchmark freeze rather than a separate access date for every provider call.
This distinction matters for hosted systems, whose implementation may change
without a public checkpoint identifier. Table~\ref{tab:appendix-models}
therefore reports the public model designations and the output property that is
relevant to cross-model scoring, rather than unverifiable backend revisions.

\begin{table}[t]
\centering
\small
\setlength{\tabcolsep}{5pt}
\caption{\textbf{Systems included in the 5,376-video evaluation.} Each model
generates one video for each of the 448 prompts. ``Access'' distinguishes
systems reached through a hosted interface from open-weight systems run
locally; ``Native audio'' indicates whether the generated output contains a
usable model-produced audio track.}
\label{tab:appendix-models}
\begin{tabular*}{\textwidth}{@{\extracolsep{\fill}}llcc@{}}
\toprule
Public model designation & Access & Videos & Native audio \\
\midrule
Seedance-2.5 & hosted & 448 & yes \\
Veo-3.1 & hosted & 448 & yes \\
Wan-3.0 & hosted & 448 & yes \\
HappyHorse-1.1 & hosted & 448 & yes \\
MiniMax-H3-Max (fal) & hosted & 448 & yes \\
Kling-3.0$^{\dagger}$ & hosted & 448 & yes \\
MiniMax-H3 & open weights, local & 448 & yes \\
LTX-2.5 & open weights, local & 448 & yes \\
Wan-2.2 & open weights, local & 448 & no \\
HunyuanVideo-1.5 & open weights, local & 448 & no \\
Kandinsky-5.0-T2V-Pro & open weights, local & 448 & no \\
LongCat-Video-13.6B & open weights, local & 448 & no \\
\bottomrule
\end{tabular*}
\par\smallskip
{\footnotesize\raggedright $^{\dagger}$Native prompt processing enabled;
generation seed fixed within each experiment.\par}
\end{table}

\subsection{Generation Protocol}

Table~\ref{tab:generation-controls} summarizes the generation controls. Each
system is run through a publicly accessible hosted interface or its released
implementation with its recommended text-to-video configuration, and every
model receives the same frozen prompt text; provider-native prompt processing, where
present, is noted per model in Table~\ref{tab:appendix-models}. The protocol
therefore fixes the textual input and the number of retained samples, but does
not claim that provider-specific samplers, guidance scales, or stochasticity
are numerically equivalent; sensitivity to the seed setting is examined for
one model in Appendix~\ref{app:seed-sensitivity}.

Resolution and frame rate are left at each provider's native output and do not
enter any score; the automated evaluator resamples all videos at 8~fps
(Appendix~\ref{app:scoring}). For models without usable native audio
(Table~\ref{tab:appendix-models}), the sound channel is excluded rather than
assigned a score of zero.

\begin{table}[t]
\centering
\small
\caption{\textbf{Generation controls and cross-model treatment.} ``Native''
means that a property is preserved from the provider output rather than
normalized by post-processing.}
\label{tab:generation-controls}
\begin{tabularx}{\textwidth}{@{}p{2.8cm}X@{}}
\toprule
Field & Protocol \\
\midrule
Prompt submission & Identical frozen English prompt for all 12 models \\
Samples per prompt & One retained generation per model--prompt pair; no
best-of-$n$ selection \\
Duration & Approximately five seconds (provider-native duration closest to the
target), continuous single-shot target; the full clip is evaluated and duration
does not enter any score \\
Resolution & Provider-recommended native output; not forcibly matched across
models \\
Frame rate & Provider-native output; automated evaluation samples at 8~fps \\
Audio & Original model-produced track retained; sound excluded when no usable
native track exists \\
Random seed & Fixed where exposed by the provider; provider default otherwise \\
Post-processing & No content-based trimming, enhancement, or candidate
selection before scoring \\
\bottomrule
\end{tabularx}
\end{table}

\subsection{Generation-Seed Sensitivity}
\label{app:seed-sensitivity}

To test whether benchmark scores reflect a reproducible model-level capability rather than an idiosyncratic realization from a single generation seed, we generated three complete 448-video batches for Kling-3.0. The model version, resolution, duration, and all other exposed generation parameters were held constant, and only the generation seed was varied across the three batches (labeled A, B, and C below). The native generation pipeline was retained. Expanded prompts were not identical across batches; therefore, this experiment measures end-to-end batch variability. Each video was evaluated once under the frozen protocol; results are reported in Tables~\ref{tab:seed-main}--\ref{tab:seed-agreement}. The leaderboard batch of Kling-3.0 was generated under a separate fixed seed and is not pooled into these statistics.

Across the three settings, the two binary metrics vary by less than 4 percentage points, whereas the emotion-conditioned Visual 9D and Full 10D macro scores vary by less than 0.4 points (Table~\ref{tab:seed-main}; the emotion-matched subsets contain 338, 325, and 341 videos). No pairwise McNemar test reaches $p<.05$ (Table~\ref{tab:seed-pairwise}). A sensitivity analysis that resamples the 16 scenario families instead of prompts yields the same conclusion. Same-prompt agreement between batches (Table~\ref{tab:seed-agreement}) is only moderate for target emotion ($\kappa\le0.58$) and substantial for the ten dimensions ($\kappa\approx0.62$), so aggregate stability coexists with substantial prompt-level variation; per-emotion ranges reach 14.3 percentage points on 56 videos. The analysis covers three seed settings of one model; it does not establish equivalence, characterize the full seed distribution, or separate generation variability from single-run evaluator noise.

\begin{table}[t]
\centering
\small
\setlength{\tabcolsep}{4pt}
\caption{\textbf{Kling-3.0 score variation across seed settings.} All batches use the same 448 prompts and exposed generation configuration; only the generation seed is changed. Visual 9D and Full 10D are emotion-conditioned macro averages on each batch's own emotion-matched subset. SD is the descriptive standard deviation over the three settings (divisor 3) and does not estimate variance over unseen seeds.}
\label{tab:seed-main}
\begin{tabular*}{\textwidth}{@{\extracolsep{\fill}}lrrrrr@{}}
\toprule
Metric & Seed A & Seed B & Seed C & Range & SD \\
\midrule
Target emotion (\%) & 75.45 & 72.54 & 76.12 & 3.57 & 1.55 \\
Emotion + target intensity (\%) & 47.77 & 45.54 & 49.33 & 3.79 & 1.56 \\
Visual 9D macro (\%) & 72.90 & 72.92 & 72.74 & 0.18 & 0.08 \\
Full 10D macro (\%) & 69.72 & 70.03 & 69.72 & 0.31 & 0.15 \\
\bottomrule
\end{tabular*}
\end{table}

\begin{table}[t]
\centering
\small
\setlength{\tabcolsep}{4pt}
\caption{\textbf{Prompt-matched pairwise differences between seed batches.} $\Delta$ is the right-hand seed minus the left-hand seed in percentage points; intervals are 95\% percentile intervals from 5,000 paired prompt bootstrap replicates; $p$ is from the two-sided exact McNemar test, applicable to the two binary metrics only.}
\label{tab:seed-pairwise}
\begin{tabular*}{\textwidth}{@{\extracolsep{\fill}}llrcc@{}}
\toprule
Comparison & Metric & $\Delta$ & 95\% CI & McNemar $p$ \\
\midrule
A $\rightarrow$ B & Target emotion & $-2.90$ & $[-6.92, +1.12]$ & .203 \\
 & Emotion + target intensity & $-2.23$ & $[-6.92, +2.23]$ & .373 \\
 & Visual 9D macro & $+0.02$ & $[-2.39, +2.53]$ & -- \\
 & Full 10D macro & $+0.31$ & $[-2.11, +2.68]$ & -- \\
\addlinespace[2pt]
A $\rightarrow$ C & Target emotion & $+0.67$ & $[-3.12, +4.46]$ & .818 \\
 & Emotion + target intensity & $+1.56$ & $[-2.90, +6.25]$ & .558 \\
 & Visual 9D macro & $-0.16$ & $[-2.69, +2.30]$ & -- \\
 & Full 10D macro & $-0.00$ & $[-2.38, +2.43]$ & -- \\
\addlinespace[2pt]
B $\rightarrow$ C & Target emotion & $+3.57$ & $[+0.00, +7.37]$ & .076 \\
 & Emotion + target intensity & $+3.79$ & $[-0.45, +8.26]$ & .111 \\
 & Visual 9D macro & $-0.18$ & $[-2.40, +2.17]$ & -- \\
 & Full 10D macro & $-0.31$ & $[-2.50, +1.86]$ & -- \\
\bottomrule
\end{tabular*}
\end{table}

\begin{table}[t]
\centering
\small
\setlength{\tabcolsep}{4pt}
\caption{\textbf{Same-prompt agreement between seed batches.} Values are agreement (\%) / Cohen's $\kappa$ between different generated videos for the same prompt, not repeated evaluations of one video.}
\label{tab:seed-agreement}
\begin{tabular*}{\textwidth}{@{\extracolsep{\fill}}lccc@{}}
\toprule
Comparison & Target emotion & Three-level intensity & Ten dimensions \\
\midrule
A vs.\ B & 80.1 / .484 & 70.1 / .527 & 81.9 / .615 \\
A vs.\ C & 83.3 / .544 & 70.5 / .532 & 82.1 / .619 \\
B vs.\ C & 83.9 / .580 & 72.1 / .560 & 82.8 / .634 \\
\bottomrule
\end{tabular*}
\end{table}

\subsection{Retention and Failure Handling}

The protocol retains weak, off-target, visually poor, or otherwise unsuccessful
generations because these are model outcomes rather than technical failures.
Only process failures---for example, an unrecoverable provider error or a file
that remains undecodable after retry---are separated from generation quality.
Transient network, provider, and decoding failures are retried under the same
frozen request settings. A sample that remains invalid is marked
\texttt{unscorable}, logged separately, and removed from the relevant
denominator; it is never converted into a negative emotion label. Safety
refusals and corrupted media are likewise logged as failure types rather than
silently replaced by a preferred candidate.

The release manifest links every valid output to its model designation, prompt
identifier, prompt hash, and media path. This one-to-one mapping supports an
audit of the expected $12\times448$ grid without exposing model identity to
either the automated evaluator or human annotators.

\section{Controlled Prompt Construction}
\label{app:prompt-construction}

\subsection{Factor Structure}

The prompt set is the complete crossing
\[
8\ \text{emotions}\times 2\ \text{scenes per emotion}\times
14\ \text{templates}\times 2\ \text{intensity levels}=448.
\]
The eight emotions are amusement, excitement, awe, contentment, anger, fear,
disgust, and sadness. Each emotion has two fixed scenes. Within a scene, every
template is instantiated once at weak intensity and once at strong intensity,
giving 224 aligned strong--weak pairs. The scenes are fixed rather than sampled,
so the benchmark supports controlled comparisons over these 16 emotion--scene
combinations and does not by itself establish scene-level generalization.

Each record is assembled from a base sentence and zero to five reusable cue
segments. The template determines which segments are written, while the
intensity member determines the preregistered expressions inside those
segments. Table~\ref{tab:appendix-templates} lists all 14 templates. ``Explicit''
denotes channels directly specified by the prompt; ``implicit'' denotes channels
left unspecified but still observed in the generated video. An implicit score
records cue co-occurrence and does not establish that the unspecified cue arose
spontaneously or was caused by another cue.

\begin{table*}[t]
\centering
\scriptsize
\setlength{\tabcolsep}{3.2pt}
\caption{\textbf{The 14 prompt templates.} Every template contributes 32
prompts: eight emotions, two scenes, and two intensity levels. ``Words'' is the
median prompt length. Visual-form segments jointly specify V1--V4; these four
channels are not independently manipulated. A dash means that no channel is
explicit.}
\label{tab:appendix-templates}
\setlength{\tabcolsep}{2.5pt}
\newcolumntype{L}[1]{>{\raggedright\arraybackslash\hsize=#1\hsize}X}
\begin{tabularx}{\textwidth}{@{}lL{1.15}lL{0.95}cccL{0.85}L{1.05}@{}}
\toprule
Template & Family & Complexity & Segments written & $|\mathrm{Exp}|$ &
$|\mathrm{atoms}|$ & Words & Explicit channels & Implicit channels \\
\midrule
EMO & emotion-label baseline & atomic & emotion label & 0 & 1 & 15 & -- & C1, C2, V1--V4, X1, X2, T1 \\
FACE & single channel & atomic & face & 1 & 1 & 26.5 & C1 & C2, V1--V4, T1 \\
BODY & single channel & atomic & body & 1 & 1 & 25.5 & C2 & C1, V1--V4, T1 \\
SCENE & single channel & atomic & scene appraisal & 1 & 1 & 24.5 & X1 & C1, C2, V1--V4, T1 \\
EVENT & single channel & atomic & event appraisal & 1 & 1 & 24.5 & X2 & C1, C2, V1--V4, T1 \\
FACE\_BODY & paired addition & composed & face + body & 2 & 2 & 42.5 & C1, C2 & V1--V4, T1 \\
SCENE\_REACT & paired addition & composed & scene + face + body & 3 & 3 & 56 & C1, C2, X1 & V1--V4, T1 \\
EVENT\_REACT & paired addition & composed & event + face + body & 3 & 3 & 57 & C1, C2, X2 & V1--V4, T1 \\
CONTEXT\_MOOD & representation & atomic & atmosphere & 4 & 1 & 18 & V1--V4 & C1, C2, T1 \\
CONTEXT\_HARD & representation & composed & cinematic form & 4 & 1 & 31 & V1--V4 & C1, C2, T1 \\
FACE\_TRANSITION & temporal probe & composed & facial transition & 1 & 1 & 39.5 & C1 & C2, V1--V4, T1 \\
ALL\_DYNAMIC & stress test & stress & scene + event + face + body + atmosphere & 8 & 5 & 79 & C1, C2, V1--V4, X1, X2 & T1 \\
EMPTY\_MOOD & no-character & composed & atmosphere & 5 & 1 & 22 & V1--V4, X1 & -- \\
EMPTY\_HARD & no-character & composed & cinematic form & 5 & 1 & 36 & V1--V4, X1 & -- \\
\bottomrule
\end{tabularx}
\end{table*}

\subsection{Explicit and Implicit Cue Matrix}

Table~\ref{tab:explicit-implicit-matrix} makes the prompt-side role assignment
explicit. The assignment is frozen before generation and hidden during scoring.
The prompt metadata covers the nine non-audio channels; temporal organization
(T1) is never written directly and is observed only in the output.

\begin{table*}[t]
\centering
\scriptsize
\setlength{\tabcolsep}{4pt}
\caption{\textbf{Prompt-side cue roles for all templates.} E = explicitly
written; I = implicit (not written but observed); -- = outside the intended
scope of that template. A1 is omitted because sound is scoring-only.}
\label{tab:explicit-implicit-matrix}
\begin{tabular*}{\textwidth}{@{\extracolsep{\fill}}lccccccccc@{}}
\toprule
Template & C1 & C2 & V1 & V2 & V3 & V4 & X1 & X2 & T1 \\
\midrule
EMO              & I & I & I & I & I & I & I & I & I \\
FACE             & E & I & I & I & I & I & -- & -- & I \\
BODY             & I & E & I & I & I & I & -- & -- & I \\
SCENE            & I & I & I & I & I & I & E & -- & I \\
EVENT            & I & I & I & I & I & I & -- & E & I \\
FACE\_BODY        & E & E & I & I & I & I & -- & -- & I \\
SCENE\_REACT      & E & E & I & I & I & I & E & -- & I \\
EVENT\_REACT      & E & E & I & I & I & I & -- & E & I \\
CONTEXT\_MOOD     & I & I & E & E & E & E & -- & -- & I \\
CONTEXT\_HARD     & I & I & E & E & E & E & -- & -- & I \\
FACE\_TRANSITION  & E & I & I & I & I & I & -- & -- & I \\
ALL\_DYNAMIC      & E & E & E & E & E & E & E & E & I \\
EMPTY\_MOOD       & -- & -- & E & E & E & E & E & -- & -- \\
EMPTY\_HARD       & -- & -- & E & E & E & E & E & -- & -- \\
\bottomrule
\end{tabular*}
\end{table*}

This matrix also defines the limits of attribution: C1, C2, X1, and X2 each
have at least one single-channel template, whereas the prompt-side effects of
V1--V4 can be attributed only to the four-channel bundle
(Table~\ref{tab:appendix-templates}), even though their output scores are
reported separately.

\FloatBarrier
\subsection{Controlled Template Pairs}
\label{app:template-pairs}

In addition to the 224 within-template intensity pairs, the design contains
seven cross-template contrasts, each holding emotion, scene, and target
intensity fixed (Table~\ref{tab:controlled-template-pairs}). Contrasts 1--6
form the pre-specified family analyzed with exact McNemar tests and Holm
correction (Appendix~\ref{app:scoring}); contrast 7 is reported descriptively
only (Table~\ref{tab:template-ei}). The replacement contrasts preserve the set
of explicit channels but not text length; they are therefore descriptive
comparisons of phrasing rather than isolated causal estimates of wording alone.

\begin{table}[t]
\centering
\small
\setlength{\tabcolsep}{4pt}
\caption{\textbf{Seven controlled template pairs.} Text not named in the
``Intended change'' column is preserved within the pair. ``Tested'' indicates
membership in the pre-specified McNemar/Holm family.}
\label{tab:controlled-template-pairs}
\begin{tabularx}{\textwidth}{@{}cllXc@{}}
\toprule
& Pair & Operation & Intended change & Tested \\
\midrule
1 & SCENE $\rightarrow$ SCENE\_REACT & add & add facial and bodily reaction to the same scene appraisal & yes \\
2 & EVENT $\rightarrow$ EVENT\_REACT & add & add facial and bodily reaction to the same event appraisal & yes \\
3 & FACE $\rightarrow$ FACE\_BODY & add & add a body segment to the same facial description & yes \\
4 & BODY $\rightarrow$ FACE\_BODY & add & add a facial segment to the same body description & yes \\
5 & CONTEXT\_MOOD $\rightarrow$ CONTEXT\_HARD & replace & experiential atmosphere $\rightarrow$ explicit cinematic form, character present & yes \\
6 & EMPTY\_MOOD $\rightarrow$ EMPTY\_HARD & replace & experiential atmosphere $\rightarrow$ explicit cinematic form, empty scene & yes \\
7 & FACE $\rightarrow$ FACE\_TRANSITION & replace & static facial state $\rightarrow$ facial change over time & no \\
\bottomrule
\end{tabularx}
\end{table}

\subsection{Representative Strong--Weak Pairs}

\begin{figure}[t]
\centering
\includegraphics[width=\textwidth]{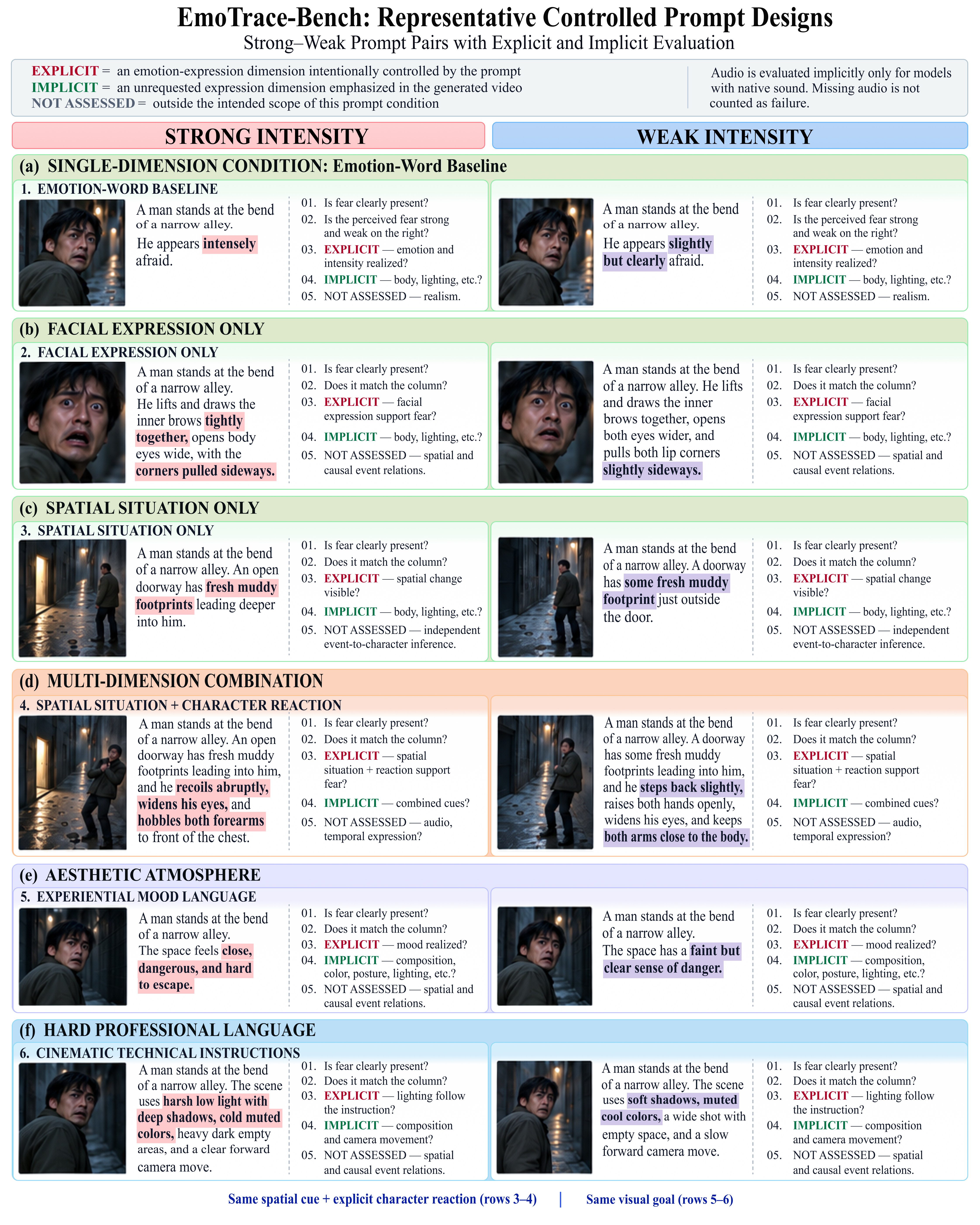}
\caption{Six prompt conditions for one fear scene, shown as strong--weak pairs
with the intensity-bearing span highlighted. Rows 1--6 correspond to the EMO,
FACE, SCENE, SCENE\_REACT, CONTEXT\_MOOD, and CONTEXT\_HARD templates
(Table~\ref{tab:appendix-templates}); rows 3--4 and rows 5--6 are contrasts 1
and 5 in Table~\ref{tab:controlled-template-pairs}.}
\label{fig:prompt-design}
\end{figure}

Figure~\ref{fig:prompt-design} shows six prompt conditions for one fear scene as
strong--weak pairs. The textual examples below are drawn from the same
amusement scene. Ellipses replace only the shared base sentence; the emphasized
span illustrates the intensity-bearing change.

\paragraph{Emotion-word baseline (EMO).}
\begin{quote}
\textbf{Weak:} A woman stands beside a fairground carousel. She appears
\emph{mildly amused}.\\
\textbf{Strong:} A woman stands beside a fairground carousel. She appears
\emph{intensely amused}.
\end{quote}

\paragraph{Facial expression (FACE).}
\begin{quote}
\textbf{Weak:} \ldots She holds a \emph{small smile} with the lips slightly
parted, both cheeks lifted, and both eyes gently narrowed.\\
\textbf{Strong:} \ldots She holds a \emph{broad, open-mouthed smile}, with both
cheeks raised high and both eyes narrowed.
\end{quote}

\paragraph{Experiential atmosphere (CONTEXT\_MOOD).}
\begin{quote}
\textbf{Weak:} \ldots The space has a \emph{light} sense of play.\\
\textbf{Strong:} \ldots The space has a \emph{clear, lively} sense of play.
\end{quote}

\paragraph{Cinematic form (CONTEXT\_HARD).}
\begin{quote}
\textbf{Weak:} \ldots The scene uses \emph{soft bright light, slightly warm
colors, a balanced medium shot, and a slow push-in}.\\
\textbf{Strong:} \ldots The scene uses \emph{bright high-key light, vivid warm
colors, a centered close shot, and a clear push-in}.
\end{quote}

\begin{figure}[t]
    \centering
    \includegraphics[width=\textwidth]{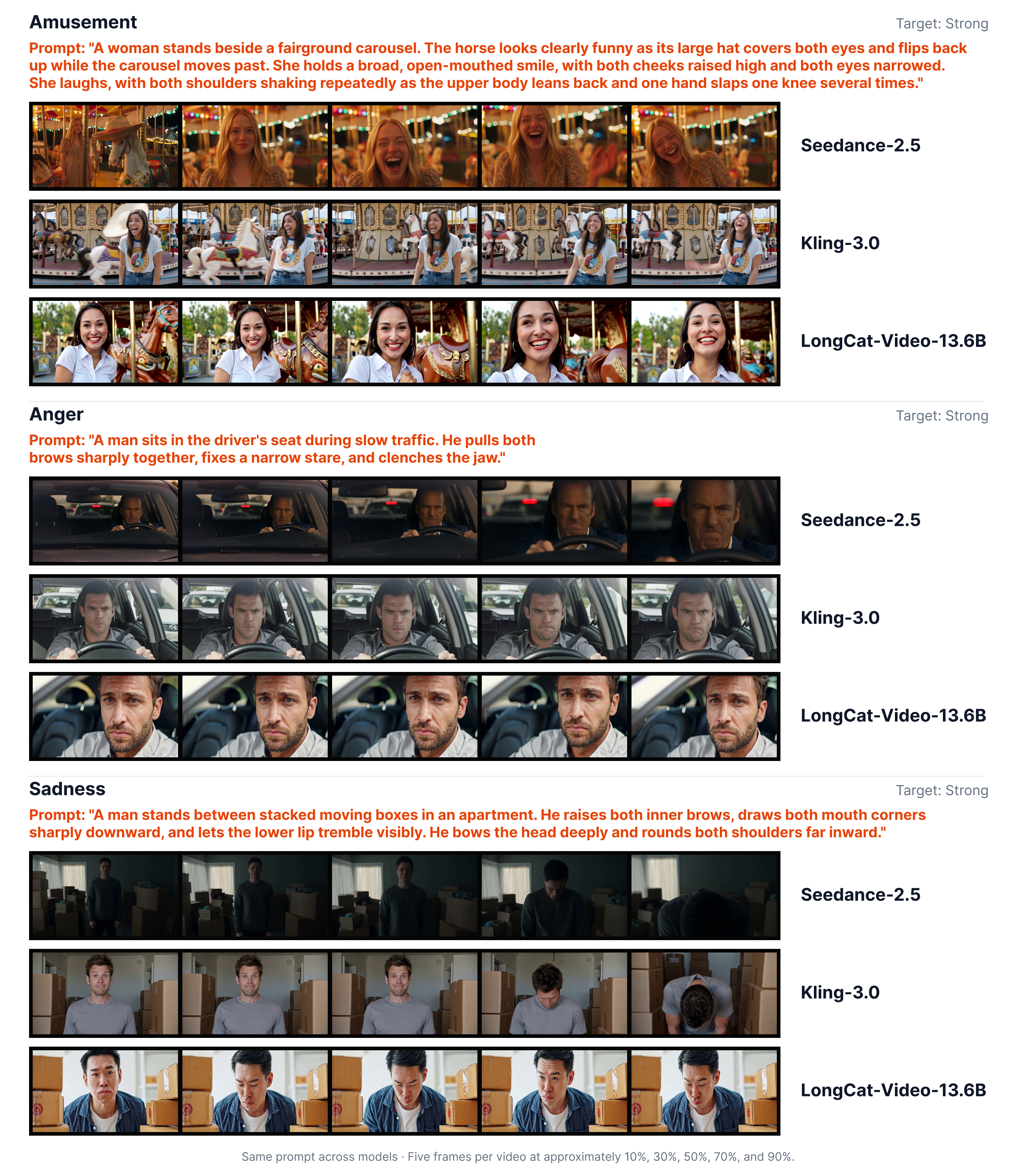}
    \caption{Frame sequences from three models for the same strong-intensity
    prompt within each of three emotions. Although every video depicts the requested scene and character,
    the realized expression ranges from clearly strong to nearly neutral, so
    emotion presence and intensity are scored separately.}
    \label{fig:video-frame-intensity}
\end{figure}

Figure~\ref{fig:video-frame-intensity} shows frame sequences from three models
for one strong-intensity prompt in each of three selected emotions; within an
emotion, all models receive the same prompt.

The first two examples vary a single intensity-bearing atom. Composite
templates change every preregistered intensity-bearing segment included by that
template. This dose is consequently not constant across templates: the
emotion-word baseline changes one atom, whereas the full stress-test template
changes five. Pair-control rates are interpreted within templates rather than
as evidence that templates with different doses are directly equivalent.

\subsection{Structural Verification}

The frozen prompt records are checked after construction rather than validated
by visual inspection alone. Each record stores its scene-pair and intensity-pair
identifiers, member role, explicit/implicit channel assignment, allowed
intensity changes, prompt text, and SHA-256 hash. The audit results are
summarized in Table~\ref{tab:prompt-audit}.

\begin{table}[t]
\centering
\small
\caption{\textbf{Structural audit of the 448-prompt set.} Structural checks
verify the frozen construction contract.}
\label{tab:prompt-audit}
\begin{tabular*}{\textwidth}{@{\extracolsep{\fill}}lr@{}}
\toprule
Check & Result \\
\midrule
Total structured prompt records & 448 \\
Strong / weak members & 224 / 224 \\
Unique prompt texts & 448 / 448 \\
Unique SHA-256 prompt hashes & 448 / 448 \\
Intensity pairs passing the pairing contract & 224 / 224 \\
Literal reuse of shared segments & 372 / 384 \\
Reuse after pronoun normalization & 384 / 384 \\
\bottomrule
\end{tabular*}
\end{table}

The 12 literal reuse exceptions (Table~\ref{tab:prompt-audit}) differ only in
pronoun disambiguation and do not alter intensity or cue content. Within
intensity pairs, the heuristic ``the longer prompt is strong'' identifies only
56.2\% of pairs, so length alone does not encode the target level; length does
differ across some cross-template contrasts (Table~\ref{tab:appendix-templates},
median words), which is why those contrasts are treated as descriptive
(Appendix~\ref{app:template-pairs}).

\subsection{Released Prompt Artifact}

The PDF reports the construction logic, role matrix, representative examples,
and audit statistics rather than reproducing all 448 texts. The project
repository (\url{https://github.com/Tonyheng6/EmoTrace-Bench}) contains the
complete prompt set and its per-record metadata,
including pair pointers, channel roles, allowed changed atoms, and hashes. This
keeps every reported comparison reproducible while avoiding a long verbatim
prompt catalogue in the paper.

\section{Scoring Protocol and Metrics}
\label{app:scoring}

\subsection{Three-Level Evaluation}

Automated evaluation and human annotation share one ordered protocol. First,
the rater decides whether the specified target emotion is present. This is a
binary verification task, not free classification among the eight emotions.
Second, if the target emotion is present, the rater classifies its perceived
intensity as weak or strong; if it is absent, intensity is recorded as
\texttt{not\_matched} rather than forced. Third, the rater independently
determines whether each of ten audiovisual channels contains directly
observable evidence supporting the target emotion.

The three outputs answer different questions and are not converted into one
another. In particular, target-emotion success and intensity are judged
directly, not inferred from the number of positive cue channels. Likewise, a
video may contain a local cue related to the target emotion without expressing
that emotion clearly at the video level.

\subsection{Operational Definition of the Ten Cue Channels}

Table~\ref{tab:appendix-cue-definitions} defines the object judged in each
channel. These are diagnostic channels
rather than interchangeable subscores of a single latent rating.

\begin{table*}[t]
\centering
\scriptsize
\setlength{\tabcolsep}{4pt}
\caption{\textbf{The ten affective cue channels, grouped by source, with their
operational definitions.} Each channel receives a binary judgment. A positive
label requires observable evidence that specifically supports the target
emotion, subject to the shared and channel-specific rules below.}
\label{tab:appendix-cue-definitions}
\setlength{\tabcolsep}{3pt}
\begin{tabularx}{\textwidth}{@{}lllXp{3.0cm}@{}}
\toprule
Group & ID & Channel & Direct object of judgment & Channel-specific prerequisite \\
\midrule
Character performance & C1 & Facial expression & Joint configuration of eyes, brows, mouth, cheeks, and jaw & An identifiable, readable face \\
 & C2 & Bodily behavior & Posture, gesture, movement direction and energy, or meaningful stillness & An identifiable, readable body \\
\addlinespace[2pt]
Visual form & V1 & Lighting & Brightness, contrast, shadow, direction, hardness, and their change over time & Visual content is readable \\
 & V2 & Color & Palette, warm--cool balance, saturation, color contrast, and change & Chromatic information is present \\
 & V3 & Composition & Placement, negative space, visual weight, balance, occlusion, and depth & Spatial configuration is visible \\
 & V4 & Visual dynamics & Camera motion, recomposition, stability, cuts, and meaningful holds & Sufficient temporal extent \\
\addlinespace[2pt]
Auditory form & A1 & Sound & Voice, nonverbal vocalization, music, ambience, sound effects, and silence contrast & A usable native audio track \\
\addlinespace[2pt]
Context & X1 & Scene / space & Openness, enclosure, scale, distance, isolation, blocking, access, and safety & Spatial relations are visible \\
 & X2 & Event / reaction & Visible trigger, affected object, reaction, and outcome & Both a trigger and its caused reaction \\
\addlinespace[2pt]
Temporal organization & T1 & Temporal organization & Persistence, intensification, transition, or development of the target emotion & More than a momentary cue \\
\bottomrule
\end{tabularx}
\end{table*}

\subsection{Shared Decision Rule}

A cue channel is labeled 1 only when all five of the following conditions hold:
\begin{enumerate}
    \item the evidence is directly observable in the video;
    \item the evidence belongs to the channel being judged;
    \item the evidence clearly supports the specified target emotion;
    \item the evidence is not an ordinary, neutral, or incidental occurrence; and
    \item the evidence can stand without borrowing evidence from another channel.
\end{enumerate}
The following cases are explicitly labeled 0: no relevant evidence; evidence
that is very weak or very brief; evidence that is occluded or unreadable;
ambiguous meaning; a conclusion available only from the hidden prompt; a
conclusion requiring an imagined backstory; evidence that depends on another
channel; evidence closer to a neutral state; or evidence closer to another
emotion.

The modality boundary is part of the rule. C1, C2, V1--V4, X1, X2, and T1 are
judged from the picture without using the soundtrack. A1 is judged from the
soundtrack without using the picture. A close-up does not by itself satisfy C1,
visible movement does not by itself satisfy C2, and a reaction without a visible
cause does not satisfy X2. T1 requires the target emotion itself to persist,
change, or develop; camera motion, color change, or an event transition alone
cannot substitute for emotional development.

Emotion-specific rule cards refine these channel-level rules. For example, a
sadness--face decision requires a readable facial configuration rather than a
single generic cue such as looking down, one tear, or a blue-gray palette. The
same visible event can therefore receive different labels for different target
emotions, but the shared five-part criterion remains fixed.

\subsection{Information Boundary and Automated Evaluation}

Both evaluation pathways can access the full video, its native audio track when
present, and the name and definition of the target emotion. Both are blinded to
the generation model, full generation prompt, requested intensity, prompt
template, explicit/implicit metadata, paired sibling video, and all other
ratings. Revealing the target emotion turns the first decision into target
verification; hiding the target intensity prevents the strong/weak decision
from copying the requested label.

Each of the 5,376 videos receives one automated evaluation pass using
Gemini~3.1 Pro with temperature 0. Videos are sampled at 8~fps, and the pass
consists of two separate calls under the frozen protocol above: the first call
returns the target-emotion decision and, when the target is present, its
perceived intensity; the second call returns the ten cue-channel judgments. No
voting or cross-call aggregation is applied to the 12-model results; the two
additional development-set runs (Appendix~\ref{app:validity}) serve only to
verify stability. Because emotion and intensity are returned together within
the first call, their separation is procedural rather than an architectural
claim that one output cannot influence the other; the cue judgments are
produced in a separate call that does not receive the first call's output.

Two bounded auxiliary modules supply measurement summaries. The visual-dynamics
module summarizes cross-frame motion for V4, and the audio module summarizes
track intensity and spectral properties for A1. Neither module receives the
target emotion, requested intensity, or target labels; neither emits a channel
label. The automated evaluator makes the final decisions from the complete
video and may not treat an auxiliary summary as a substitute for observable
emotional evidence.

\subsection{Metric Definitions}

Let $P_m$ be the valid videos generated by model $m$; the reported 12-model
experiment has $|P_m|=448$ for every model. Let $M_{mp}\in\{0,1\}$ indicate
target-emotion success, $I_{mp}\in\{\mathrm{weak},\mathrm{strong}\}$ denote
perceived intensity when $M_{mp}=1$, and $T_p$ denote the requested intensity.
Let $Y_{mpd}\in\{0,1\}$ be the score for cue channel $d$. Percentages below are
multiplied by 100 in the reported tables.

\paragraph{Target emotion (Emotion).}
\[
\mathrm{Emotion}(m)=
\frac{\sum_{p\in P_m} M_{mp}}{|P_m|}.
\]

\paragraph{Joint emotion and intensity (E+I).}
\[
\mathrm{E{+}I}(m)=
\frac{\sum_{p\in P_m}
\mathbf{1}\!\left[M_{mp}=1\ \land\ I_{mp}=T_p\right]}{|P_m|}.
\]
This is an unconditional conjunctive success rate: a video cannot receive
credit for matching the requested degree when the target emotion is absent.

\paragraph{Per-channel expression depth.}
For a valid channel $d$, cue support is measured only among videos in which the
target emotion succeeds:
\[
Q_d(m)=
\frac{\sum_{p\in P_m} M_{mp}Y_{mpd}}
{\sum_{p\in P_m} M_{mp}}.
\]
Thus, cue depth describes how successful emotional outputs are expressed; it
does not reward a model for failing to generate the target emotion. It must be
read alongside Emotion because models with different success rates contribute
different conditional subsets.

\paragraph{Visual 9D and Full 10D.}
Let
\[
\mathcal{D}_9=\{\mathrm{C1,C2,V1,V2,V3,V4,X1,X2,T1}\},
\qquad
\mathcal{D}_{10}=\mathcal{D}_9\cup\{\mathrm{A1}\}.
\]
Then
\[
\mathrm{Visual\ 9D}(m)=\frac{1}{9}\sum_{d\in\mathcal{D}_9}Q_d(m),
\qquad
\mathrm{Full\ 10D}(m)=\frac{1}{10}\sum_{d\in\mathcal{D}_{10}}Q_d(m).
\]
Full 10D is reported only for models with usable native audio
(Table~\ref{tab:appendix-models}); silent models receive ``--'', not zero,
for A1 and Full 10D.

\paragraph{Paired intensity control.}
For each of the 224 aligned pairs $k$, let $p_k^{-}$ be the weak-target member
and $p_k^{+}$ the strong-target member. A pair succeeds only when both emotions
are present and both perceived levels are correct:
\[
\mathrm{PairSuccess}_{mk}=
\mathbf{1}\!\left[
M_{m p_k^-}=1 \land I_{m p_k^-}=\mathrm{weak}
\land M_{m p_k^+}=1 \land I_{m p_k^+}=\mathrm{strong}
\right],
\]
\[
\mathrm{PairedControl}(m)=
\frac{1}{224}\sum_{k=1}^{224}\mathrm{PairSuccess}_{mk}.
\]
For failure analysis, pairs are additionally partitioned into correctly
ordered, collapsed to the same perceived level, reversed, or not comparable
because at least one member fails the target emotion.

\subsection{Exact Denominators and Missing Values}

Table~\ref{tab:metric-denominators} collects the denominator rules used for all
reported metrics. \texttt{Unscorable} denotes a process failure after retry and
is never converted to a negative model outcome. A silent output is different:
its visual and emotion labels remain valid, but it has no object for A1 and is
excluded from audio-dependent aggregates.

\begin{table}[t]
\centering
\small
\setlength{\tabcolsep}{4pt}
\caption{\textbf{Exact metric denominators.}}
\label{tab:metric-denominators}
\begin{tabularx}{\textwidth}{@{}p{2.8cm}p{4.9cm}X@{}}
\toprule
Metric & Denominator & Treatment of missing or inapplicable values \\
\midrule
Emotion & all valid videos ($448$ per model) & exclude and separately count
\texttt{unscorable} process failures \\
E+I & all valid videos ($448$ per model) & emotion failure receives no joint
credit; intensity is not forced \\
Cue channel $d$ & videos with $M=1$ & exclude a channel only when no valid
object exists, such as A1 without native audio \\
Visual 9D & mean of nine visual/context/temporal channel rates & always reported
for a valid model run \\
Full 10D & mean of ten channel rates & not reported for silent models \\
Paired control & all 224 strong--weak pairs & a pair with either emotion absent
is a failed pair, not an intensity reversal \\
\bottomrule
\end{tabularx}
\end{table}

For later human analyses, an intensity vote is meaningful only after that
annotator marks the target emotion present. A human consensus tie is recorded
as \texttt{undecided} and excluded from the corresponding agreement
denominator; an unavailable channel is recorded as not applicable rather than
as zero. These exclusions are reported with their valid counts so that coverage
and agreement are not conflated.

\subsection{Bootstrap, McNemar Tests, and Multiplicity}

Uncertainty in target-emotion performance is estimated with 5,000
prompt-level paired bootstrap resamples. The same resampled prompt indices are
applied to every model, preserving the matched benchmark design. The six
reported Emotion tiers are based on 95\% bootstrap intervals for paired
differences. Models within a tier are not ordered, and tier assignments on a
bootstrap boundary are marked as seed-sensitive.

The six pre-specified template contrasts
(Table~\ref{tab:controlled-template-pairs}, contrasts 1--6) are analyzed
separately for each model. Each condition contains 32 videos. The binary outcome is joint E+I
success, and the two matched conditions are compared with an exact paired
McNemar test. The six planned tests within a model form one family and are
adjusted with Holm's step-down procedure at $\alpha=0.05$. Effect sizes and
confidence intervals are reported alongside adjusted significance. Given the
small cells, a nonsignificant contrast is treated as inconclusive rather than
as evidence that the two prompt forms are equivalent.

\section{Human Reference and Automated-Evaluation Validity}
\label{app:validity}

This section reports construction of the human consensus, agreement among trained annotators, automated-evaluator agreement with the first-pass and reviewed human reference, and per-dimension performance on the test set. We report results against both the first-pass and the reviewed reference, together with their valid counts.

\subsection{Human Reference and Consensus Construction}

\begin{table}[t]
\centering
\small
\setlength{\tabcolsep}{4pt}
\caption{\textbf{Agreement among human judgments.} Development-set values summarize five trained annotators. Test-set agreement is computed only between the two full-coverage annotators; the third, senior annotator serves only as a reviewer.}
\label{tab:human-reference}
\begin{tabular*}{\textwidth}{@{\extracolsep{\fill}}lccc@{}}
\toprule
Reference & Target emotion & Intensity & Ten dimensions \\
\midrule
Development: pairwise agreement (\%) & 85.7 & 77.9 & 80.8 \\
Development: Fleiss' $\kappa$ & .65 & .50 & .60 \\
Reviewed reference vs.\ trained majority (\%) & 90.4 & 90.9 & 94.2 \\
Test: two-annotator agreement (\%) & 81.7 & 75.5 & 80.2 \\
Test: Cohen's $\kappa$ & .54 & .50 & .58 \\
\bottomrule
\end{tabular*}
\end{table}

\begin{table}[t]
\centering
\small
\setlength{\tabcolsep}{5pt}
\caption{\textbf{First-pass consensus and second-review statistics.} Counts refer to fields.}
\label{tab:consensus-construction}
\begin{tabularx}{\textwidth}{@{}Xrr@{}}
\toprule
Item & Development & Test \\
\midrule
First-pass agreed fields & 8,899 & 1,780 \\
First-pass disputed fields & 317 & 524 \\
Tie-broken by trained annotators & 206 & -- \\
Excluded from first-pass reference (no consensus) & 111 & 524 \\
Fields changed at second review & 308 & 297 \\
Second-review annotators & same two senior annotators & a third, senior annotator \\
\bottomrule
\end{tabularx}
\end{table}

Table~\ref{tab:consensus-construction} summarizes consensus construction. On the development set, first-pass disagreements between the two senior annotators were resolved by the majority of available labels from five trained annotators, and fields without consensus were excluded. On the test set, agreements between the two full-coverage annotators formed the first-pass consensus; disputed fields had no tie-break and were excluded from the first-pass reference. After the automated evaluator was frozen, a second review was conducted: on the development set by the same two senior annotators, on the test set by a third, senior annotator. Fields without a reviewed label are excluded from analyses against the reviewed reference. Reviewers had no access to automated outputs. The test set was generated by a generator not represented in the development set and was not used in protocol development or calibration; it was scored once under the frozen protocol.

\subsection{First-Pass and Reviewed Human Reference}

\begin{table*}[t]
\centering
\scriptsize
\setlength{\tabcolsep}{4pt}
\caption{\textbf{Complete agreement results between the automated evaluator and the human reference.} Values are agreement (\%) / Cohen's $\kappa$; development-set values are mean $\pm$ standard deviation over three independent runs (the 12-model run plus two stability runs); test-set values are from the single run. Parentheses give valid counts.}
\label{tab:validity-full}
\resizebox{\textwidth}{!}{%
\begin{tabular}{@{}lcccc@{}}
\toprule
Reference & Target emotion & Conditional intensity & Joint three-way & Ten dimensions \\
\midrule
Development, first-pass human reference
& $88.1{\pm}0.7$ / $.670{\pm}.021$
& $79.1{\pm}0.1$ / $.578{\pm}.003$ (637)
& $73.4{\pm}0.6$ / $.588{\pm}.009$
& $83.0{\pm}0.2$ / $.625{\pm}.003$ (7,424) \\
Test, first-pass human reference
& 84.7 / .589 (183)
& 88.3 / .764 (94)
& 77.0 / .652
& 80.5 / .585 (1,489) \\
Development, reviewed human reference
& $89.9{\pm}0.9$ / $.716{\pm}.024$
& $93.0{\pm}1.3$ / $.858{\pm}.026$ (641)
& $84.8{\pm}1.8$ / $.764{\pm}.028$
& $85.1{\pm}0.3$ / $.669{\pm}.007$ (7,424) \\
Test, reviewed human reference
& 90.2 / .762 (224)
& 92.0 / .840 (125)
& 87.1 / .806
& 83.0 / .643 (1,639) \\
\bottomrule
\end{tabular}
}
\vspace{2pt}

\parbox{\textwidth}{\scriptsize
Column definitions follow Table~\ref{tab:judge-validity}.}
\end{table*}

Relative to the development set, test-set agreement against the reviewed reference differs by $+0.3$, $-1.0$, and $-2.1$ percentage points for target emotion, conditional intensity, and the ten dimensions; video-cluster bootstrap 95\% intervals for these differences are $[-4.20, +4.50]$, $[-6.47, +3.87]$, and $[-4.87, +0.47]$, all including zero. On the 183 test videos present in both references, emotion agreement rises from 84.7\% to 90.7\% after relabeling, and the 41 videos without a first-pass consensus reach 87.8\%, so the reviewed-reference figure reflects both label revision and denominator change.

\subsection{Per-Dimension Results}

\begin{table*}[t]
\centering
\scriptsize
\setlength{\tabcolsep}{3.5pt}
\caption{\textbf{Per-dimension agreement against the reviewed human reference.} Each cell reports agreement (\%), Cohen's $\kappa$, valid count, and the human-to-automated pass rates; run averaging follows Table~\ref{tab:validity-full}.}
\label{tab:validity-dimensions}
\resizebox{\textwidth}{!}{%
\begin{tabular}{@{}lcc@{}}
\toprule
Dimension & Development & Test \\
\midrule
Facial expression & 87.9 / .74 / 768 / 64$\rightarrow$63 & 84.6 / .69 / 156 / 58$\rightarrow$52 \\
Bodily behavior & 85.1 / .70 / 768 / 53$\rightarrow$59 & 84.1 / .68 / 157 / 51$\rightarrow$52 \\
Lighting & 82.4 / .55 / 896 / 74$\rightarrow$72 & 77.7 / .50 / 224 / 70$\rightarrow$63 \\
Color & 85.3 / .64 / 896 / 71$\rightarrow$71 & 83.5 / .64 / 224 / 67$\rightarrow$63 \\
Composition & 84.9 / .55 / 896 / 80$\rightarrow$77 & 79.3 / .30 / 174 / 93$\rightarrow$74 \\
Visual dynamics & 84.0 / .64 / 896 / 64$\rightarrow$69 & 86.0 / .72 / 186 / 49$\rightarrow$48 \\
Sound & 83.3 / .66 / 896 / 40$\rightarrow$46 & 85.6 / .71 / 195 / 49$\rightarrow$52 \\
Scene / space & 84.0 / .52 / 384 / 78$\rightarrow$80 & 87.5 / .65 / 96 / 80$\rightarrow$74 \\
Event / reaction & 81.9 / .63 / 256 / 45$\rightarrow$42 & 79.7 / .58 / 64 / 45$\rightarrow$28 \\
Temporal organization & 90.8 / .76 / 768 / 74$\rightarrow$74 & 82.8 / .62 / 163 / 72$\rightarrow$62 \\
\bottomrule
\end{tabular}
}
\end{table*}

On the development set, seven dimensions have $\kappa\geq0.60$ (Table~\ref{tab:validity-dimensions}). Lighting, composition, and scene / space have $\kappa$ between 0.52 and 0.55; each also has a human pass rate of at least 74\%, indicating sensitivity to prevalence. On the test set, composition is the only dimension below $\kappa=0.50$: a 93\% human pass rate and 74\% automated pass rate yield 79.3\% agreement but $\kappa=0.30$. Automated pass rates are lower than human pass rates on most test-set dimensions, whereas the two are broadly similar on the development set.

\subsection{Per-Emotion Observation}

Per-emotion analysis shows lower human agreement on intensity for low-arousal emotions: pairwise agreement is 70.1\% for contentment and 73.4\% for awe. This pattern is consistent with lower overall human agreement on intensity than on target emotion or the ten-dimensional judgments, indicating that the strong--weak boundary is intrinsically more subjective.

\subsection{Judge-Backbone Sensitivity}
\label{app:judge-backbone}

To assess dependence on a single evaluator, we applied the frozen protocol to the 224 test-set videos with an independent Qwen3.8-Omni-Flash backbone (5~fps, temperature 0, high reasoning) and re-ran Gemini~3.1 Pro at a matched 5~fps as a control. Both evaluators used the same videos, rubric, auxiliary summaries, and two-stage structure, with identical rule-file hashes; the only Qwen-specific change was a non-semantic output-format adaptation. Qwen received no human labels, and all 224 videos and their 448 stage calls (two per video) completed without failure. Table~\ref{tab:backbone-human} reports agreement with the human reference and Table~\ref{tab:backbone-direct} the direct agreement between evaluators. Against the reviewed reference, Qwen's target-emotion $\kappa$ is 0.627 (95\% CI $[0.508, 0.735]$) and the Qwen$-$Gemini difference is $-0.095$ (95\% CI $[-0.212, 0.025]$), so the two are not statistically distinguishable, nor shown equivalent. Qwen's sound $\kappa$ against the reviewed reference is 0.631 (81.5\%), and it correctly separated 7 of 8 audio-present versus audio-removed control pairs. Because the pilot covers one generator, it does not constitute a cross-evaluator replication of the 12-model ranking.

\begin{table}[t]
\centering
\small
\setlength{\tabcolsep}{5pt}
\caption{\textbf{Test-set agreement of two evaluator backbones with the human reference.} Values are agreement (\%) / Cohen's $\kappa$; both evaluators run at 5~fps under the frozen protocol.}
\label{tab:backbone-human}
\begin{tabular*}{\textwidth}{@{\extracolsep{\fill}}lcc@{}}
\toprule
Reference / task & Qwen3.8-Omni-Flash & Gemini~3.1 Pro \\
\midrule
First-pass: target emotion & 81.4 / .509 & 83.6 / .574 \\
First-pass: joint three-way & 70.3 / .542 & 73.6 / .602 \\
First-pass: conditional intensity & 84.3 / .679 & 82.8 / .653 \\
First-pass: ten dimensions & 83.1 / .627 & 79.9 / .575 \\
\addlinespace[2pt]
Reviewed: target emotion & 84.4 / .627 & 88.4 / .721 \\
Reviewed: joint three-way & 74.2 / .610 & 80.4 / .705 \\
Reviewed: conditional intensity & 81.2 / .629 & 85.1 / .702 \\
Reviewed: ten dimensions & 82.2 / .611 & 81.3 / .610 \\
\bottomrule
\end{tabular*}
\end{table}

\begin{table}[t]
\centering
\small
\setlength{\tabcolsep}{3pt}
\caption{\textbf{Direct agreement between evaluator runs on the same test-set videos.} Values are Cohen's $\kappa$. The last row compares the 8~fps setting used in the main results with the 5~fps control.}
\label{tab:backbone-direct}
\begin{tabular*}{\textwidth}{@{\extracolsep{\fill}}lccccc@{}}
\toprule
Pair & Emotion & Joint three-way & Ten dimensions & Visual 9D & Sound \\
\midrule
Qwen3.8-Omni-Flash vs.\ Gemini (5~fps) & .640 & .622 & .678 & .676 & .678 \\
Qwen3.8-Omni-Flash vs.\ Gemini (8~fps) & .679 & .630 & .680 & .682 & .660 \\
Gemini 8~fps vs.\ 5~fps & .875 & .844 & .861 & -- & .893 \\
\bottomrule
\end{tabular*}
\end{table}

\section{Additional Results}
\label{app:additional-results}

This appendix reports the numeric values behind the cue-level figures in the
main text and two per-model breakdowns that the main text summarizes only in
aggregate: paired intensity control and joint emotion-and-intensity accuracy
for each of the 14 prompt templates. All values come from the same single
blinded automated evaluation per video as the main results. Rows follow the
ordering of Figure~\ref{fig:model-overall}: audio-capable models first, silent
models second, and within each block by target-emotion accuracy.

\subsection{Cue-Level Expression Profiles}

Table~\ref{tab:ten-model-cues} gives the values plotted in the left panel of
Figure~\ref{fig:implicit-cue-profiles}.

\begin{table*}[!htbp]
\centering
\scriptsize
\setlength{\tabcolsep}{3.8pt}
\renewcommand{\arraystretch}{1.12}
\caption{\textbf{Cue-level expression profiles of 12 T2V models.} Each score
is the percentage of target-emotion-successful videos in which the corresponding
channel supports the target emotion. $N$ is the number of such videos for each
model.}
\label{tab:ten-model-cues}
\resizebox{\textwidth}{!}{%
\begin{tabular}{@{}l r cc cccc c cc c@{}}
\toprule
& & \multicolumn{2}{c}{\textit{Character}}
& \multicolumn{4}{c}{\textit{Visual form}}
& \multicolumn{1}{c}{\textit{Audio}}
& \multicolumn{2}{c}{\textit{Context}}
& \multicolumn{1}{c}{\textit{Temporal}} \\
\cmidrule(lr){3-4}\cmidrule(lr){5-8}\cmidrule(lr){9-9}
\cmidrule(lr){10-11}\cmidrule(lr){12-12}
Model & $N$
& C1 & C2
& V1 & V2 & V3 & V4
& A1
& X1 & X2
& T1 \\
\midrule
Seedance-2.5 & 390
& \textbf{86.4} & 70.1
& \textbf{85.1} & \textbf{84.9} & \textbf{89.7} & \textbf{93.6}
& \textbf{82.6}
& 89.8 & 58.6
& \textbf{93.2} \\
Veo-3.1 & 380
& 83.6 & 74.1
& 76.8 & 79.0 & 87.4 & 44.5
& 61.0
& \textbf{91.7} & \textbf{62.3}
& 91.4 \\
Wan-3.0 & 358
& 82.2 & 68.1
& 76.5 & 79.6 & 85.8 & 33.8
& 78.5
& 91.6 & 46.4
& 87.8 \\
HappyHorse-1.1 & 357
& 80.7 & 73.1
& 72.8 & 76.2 & 82.6 & 86.8
& 75.3
& 88.8 & 51.9
& 85.9 \\
MiniMax-H3-Max (fal) & 333
& 81.4 & 66.4
& 74.2 & 80.8 & 85.0 & 33.0
& 52.2
& 89.0 & 49.0
& 88.9 \\
Kling-3.0 & 328
& 77.1 & \textbf{75.0}
& 69.8 & 69.8 & 82.3 & 61.3
& 41.2
& 90.1 & 52.0
& 89.6 \\
MiniMax-H3 & 304
& 75.3 & 68.6
& 70.4 & 71.7 & 77.3 & 48.0
& 65.5
& 80.8 & 38.3
& 84.7 \\
LTX-2.5 & 301
& 81.6 & 70.6
& 74.1 & 68.8 & 76.7 & 56.1
& 42.5
& 81.4 & 33.0
& 86.3 \\
\addlinespace[2pt]
Wan-2.2 & 302
& 74.7 & 63.8
& 67.5 & 71.8 & 78.2 & 21.2
& --
& 82.3 & 27.1
& 87.5 \\
HunyuanVideo-1.5 & 249
& 81.7 & 63.9
& 57.8 & 62.2 & 64.7 & 47.8
& --
& 78.5 & 21.2
& 86.8 \\
Kandinsky-5.0-T2V-Pro & 226
& 75.2 & 61.1
& 65.0 & 69.9 & 70.3 & 62.8
& --
& 80.2 & 19.7
& 83.3 \\
LongCat-Video-13.6B & 221
& 75.0 & 57.2
& 76.5 & 77.8 & 77.4 & 41.2
& --
& 81.1 & 17.5
& 82.2 \\
\midrule
\textit{Mean} &
& \textit{79.6} & \textit{67.7}
& \textit{72.2} & \textit{74.4} & \textit{79.8} & \textit{52.5}
& \textit{62.4}
& \textit{85.4} & \textit{39.7}
& \textit{87.3} \\
\textit{Range} &
& \textit{11.7} & \textit{17.8}
& \textit{27.3} & \textit{22.6} & \textit{25.1} & \textit{72.4}
& \textit{41.4}
& \textit{13.3} & \textit{44.8}
& \textit{11.0} \\
\bottomrule
\end{tabular}
}
\vspace{2pt}

\parbox{\textwidth}{\scriptsize
Channel abbreviations follow Table~\ref{tab:appendix-cue-definitions}. A1 is
unavailable for the four silent models (--); its mean and range are computed
over the eight audio-capable models. Bold denotes the largest observed value in each column
and does not imply statistical significance. All ten channels have a
cross-model range of at least 10 percentage points, but T1 is saturated (at
least 80\% for every model) and the C1 and T1 ranges only just exceed 10
points, so differences within these two columns are not interpreted.}
\end{table*}

\subsection{Implicit Cue Expression}

Table~\ref{tab:implicit-cues} gives the values plotted in the right panel of
Figure~\ref{fig:implicit-cue-profiles}. The bottom rows contrast the
12-model mean support under unspecified and specified conditions; the
difference indicates how strongly each channel depends on being written in the
prompt. These are co-occurrence rates and do not establish that an unspecified
cue was caused by another cue.

\begin{table*}[!htbp]
\centering
\scriptsize
\setlength{\tabcolsep}{3.8pt}
\renewcommand{\arraystretch}{1.12}
\caption{\textbf{Implicit cue-expression profiles of 12 T2V models.} Each
score is the percentage of target-emotion-successful videos, among those whose
prompt does not explicitly specify the channel, in which that channel supports
the target emotion. Implicit 9D is the mean over the nine visual, contextual,
and temporal channels.}
\label{tab:implicit-cues}
\resizebox{\textwidth}{!}{%
\begin{tabular}{@{}l cc cccc c cc c r@{}}
\toprule
& \multicolumn{2}{c}{\textit{Character}}
& \multicolumn{4}{c}{\textit{Visual form}}
& \multicolumn{1}{c}{\textit{Audio}}
& \multicolumn{2}{c}{\textit{Context}}
& \multicolumn{1}{c}{\textit{Temporal}} & \\
\cmidrule(lr){2-3}\cmidrule(lr){4-7}\cmidrule(lr){8-8}
\cmidrule(lr){9-10}\cmidrule(lr){11-11}
Model
& C1 & C2
& V1 & V2 & V3 & V4
& A1$^{\P}$
& X1$^{\S}$ & X2$^{\S}$
& T1$^{\P}$ & Implicit 9D \\
\midrule
Seedance-2.5
& \textbf{71.5} & 54.9
& \textbf{81.2} & \textbf{81.6} & \textbf{88.1} & \textbf{94.6}
& \textbf{82.6}
& 90.3 & 29.0
& \textbf{93.2} & 76.1 \\
Veo-3.1
& 67.4 & 57.6
& 71.0 & 73.0 & 85.3 & 36.9
& 61.0
& 93.5 & 35.5
& 91.4 & 68.0 \\
Wan-3.0
& 63.5 & 48.5
& 70.3 & 74.2 & 81.8 & 25.0
& 78.5
& 90.0 & 6.7
& 87.8 & 60.9 \\
HappyHorse-1.1
& 70.2 & \textbf{61.1}
& 66.8 & 70.2 & 79.0 & 86.5
& 75.3
& 83.3 & 36.7
& 85.9 & 71.1 \\
MiniMax-H3-Max (fal)
& 61.3 & 42.5
& 68.3 & 74.9 & 81.5 & 23.8
& 52.2
& 78.6 & 17.9
& 88.9 & 59.7 \\
Kling-3.0
& 57.9 & 55.0
& 64.4 & 63.6 & 78.7 & 54.7
& 41.2
& 96.8 & 32.3
& 89.6 & 65.9 \\
MiniMax-H3
& 46.8 & 52.0
& 64.5 & 63.5 & 73.0 & 38.0
& 65.5
& 75.0 & 14.3
& 84.7 & 56.9 \\
LTX-2.5
& 60.2 & 49.1
& 66.0 & 61.6 & 71.8 & 46.1
& 42.5
& 69.0 & 17.2
& 86.3 & 58.6 \\
\addlinespace[2pt]
Wan-2.2
& 52.1 & 40.9
& 62.1 & 64.0 & 75.4 & 16.3
& --
& 84.0 & 8.0
& 87.5 & 54.5 \\
HunyuanVideo-1.5
& 67.3 & 43.6
& 55.8 & 55.8 & 60.9 & 40.8
& --
& 76.2 & 0.0
& 86.8 & 54.1 \\
Kandinsky-5.0-T2V-Pro
& 48.4 & 43.9
& 61.2 & 66.9 & 69.4 & 59.4
& --
& 80.0 & 15.0
& 83.3 & 58.6 \\
LongCat-Video-13.6B
& 61.3 & 36.5
& 75.2 & 74.5 & 77.2 & 33.1
& --
& 66.7 & 13.3
& 82.2 & 57.8 \\
\midrule
\textit{Mean, channel unspecified}
& \textit{60.7} & \textit{48.8}
& \textit{67.2} & \textit{68.6} & \textit{76.8} & \textit{46.3}
& \textit{62.4}
& \textit{81.9} & \textit{18.8}
& \textit{87.3} & \textit{61.8} \\
\textit{Mean, channel specified}
& \textit{90.8} & \textit{86.3}
& \textit{82.6} & \textit{86.3} & \textit{85.9} & \textit{65.5}
& --
& \textit{86.2} & \textit{47.8}
& -- & \\
\textit{Specified $-$ unspecified (pp)}
& \textit{+30.1} & \textit{+37.5}
& \textit{+15.3} & \textit{+17.7} & \textit{+9.1} & \textit{+19.2}
& --
& \textit{+4.2} & \textit{+29.0}
& -- & \\
\bottomrule
\end{tabular}
}
\vspace{2pt}

\parbox{\textwidth}{\scriptsize
Channel abbreviations follow Table~\ref{tab:appendix-cue-definitions}; pp = percentage points.
$^{\P}$A1 and T1 are never specified in prompts, so all of their support is
implicit and no ``specified'' counterpart exists; A1 is unavailable for the four
silent models and its mean is computed over the eight audio-capable models.
$^{\S}$X1 and X2 are unspecified in only about 30 videos per model
(confidence intervals of roughly $\pm 15$ percentage points), so these two columns are not
bolded. Bold denotes the largest observed value in each remaining column and
does not imply statistical significance. Channels whose cross-model range is
below 10 percentage points are not interpreted.}
\end{table*}

\subsection{Paired Intensity Control}

Table~\ref{tab:intensity-control} decomposes the 224 aligned strong--weak
prompt pairs per model into four mutually exclusive outcomes. A pair is
\emph{correctly ordered} when the weak member is judged weak and the strong
member strong; \emph{collapsed} when both members receive the same perceived
intensity; \emph{reversed} when the strong member is judged weaker than the
weak member; and \emph{not comparable} when at least one member fails the
target emotion. The dominant failure for every model is collapse rather than
reversal.

\begin{table}[!htbp]
\centering
\small
\setlength{\tabcolsep}{5pt}
\caption{\textbf{Outcomes of the 224 strong--weak prompt pairs per model.} All
values are percentages of pairs and sum to 100 within a row.}
\label{tab:intensity-control}
\begin{tabular*}{\textwidth}{@{\extracolsep{\fill}}lrrrr@{}}
\toprule
Model & Correctly ordered & Collapsed & Reversed & Not comparable \\
\midrule
Seedance-2.5 & 27.7 & 47.3 & 5.8 & 19.2 \\
Veo-3.1 & 28.6 & 48.2 & 1.3 & 21.9 \\
Wan-3.0 & 26.8 & 43.8 & 0.9 & 28.6 \\
HappyHorse-1.1 & \textbf{29.5} & 36.6 & 3.1 & 30.8 \\
MiniMax-H3-Max (fal) & 23.2 & 36.6 & 1.3 & 38.8 \\
Kling-3.0 & 19.6 & 39.7 & 1.3 & 39.3 \\
MiniMax-H3 & 19.6 & 32.1 & 1.8 & 46.4 \\
LTX-2.5 & 27.2 & 25.9 & 1.3 & 45.5 \\
\addlinespace[2pt]
Wan-2.2 & 17.9 & 36.6 & 1.3 & 44.2 \\
HunyuanVideo-1.5 & 21.9 & 20.5 & 0.9 & 56.7 \\
Kandinsky-5.0-T2V-Pro & 14.3 & 17.9 & 0.9 & 67.0 \\
LongCat-Video-13.6B & 10.3 & 22.3 & 1.3 & 66.1 \\
\midrule
\textit{Mean} & \textit{22.2} & \textit{34.0} & \textit{1.8} & \textit{42.0} \\
\bottomrule
\end{tabular*}
\vspace{2pt}

\parbox{\textwidth}{\scriptsize
Differences between adjacent models in the first column lie within sampling
uncertainty (roughly $\pm 6$ percentage points), so this table does not rank models on
intensity control; bold marks the largest observed value only. The
``not comparable'' share follows target-emotion accuracy: the lower a model's
Emotion score, the fewer pairs enter the intensity comparison.}
\end{table}

\subsection{Per-Template Joint Accuracy}

Table~\ref{tab:template-ei} reports joint emotion-and-intensity accuracy (E+I)
for each of the 14 prompt templates and each model, together with the paired
differences for the six pre-specified contrasts
(Table~\ref{tab:controlled-template-pairs}); the two ``experiential
$\rightarrow$ cinematic'' $\Delta$ rows are the character-present and
empty-scene instances of the same replacement. The FACE\_TRANSITION row is
included for completeness; contrast 7 was not tested, and no significance is
claimed for its difference from FACE.

\begin{table*}[!htbp]
\centering
\scriptsize
\setlength{\tabcolsep}{3pt}
\renewcommand{\arraystretch}{1.1}
\caption{\textbf{Joint emotion-and-intensity accuracy (\%) by prompt template
and model.} Regular rows report E+I over the 32 videos of each template.
Italic $\Delta$ rows report the paired difference (in percentage points) of the
row immediately above from the named template, for the six pre-specified
contrasts; $^{*}$ denotes $p<0.05$ under an exact paired McNemar test after
model-wise Holm correction over the six contrasts. ``Mean'' is the
12-model mean; ``Sig.'' counts models with a significant difference.}
\label{tab:template-ei}
\resizebox{\textwidth}{!}{%
\begin{tabular}{@{}l rrrrrrrrrrrr rc@{}}
\toprule
Template
& \rotatebox{90}{Seedance-2.5}
& \rotatebox{90}{Veo-3.1}
& \rotatebox{90}{Wan-3.0}
& \rotatebox{90}{HappyHorse-1.1}
& \rotatebox{90}{MiniMax-H3-Max (fal)}
& \rotatebox{90}{Kling-3.0}
& \rotatebox{90}{MiniMax-H3}
& \rotatebox{90}{LTX-2.5}
& \rotatebox{90}{Wan-2.2}
& \rotatebox{90}{HunyuanVideo-1.5}
& \rotatebox{90}{Kandinsky-5.0-T2V-Pro}
& \rotatebox{90}{LongCat-Video-13.6B}
& \rotatebox{90}{Mean}
& \rotatebox{90}{Sig.} \\
\midrule
\multicolumn{15}{@{}l}{\textit{Emotion-label baseline}} \\
EMO & 66 & 72 & 81 & 75 & 72 & 62 & 62 & 69 & 62 & 50 & 44 & 25 & 62 & \\
\addlinespace[3pt]
\multicolumn{15}{@{}l}{\textit{Character performance}} \\
FACE & 69 & 81 & 50 & 62 & 69 & 66 & 56 & 53 & 47 & 53 & 34 & 41 & 57 & \\
BODY & 56 & 41 & 47 & 53 & 38 & 47 & 25 & 50 & 28 & 28 & 38 & 25 & 40 & \\
FACE\_BODY & 69 & 69 & 66 & 81 & 62 & 56 & 66 & 78 & 50 & 62 & 66 & 41 & 64 & \\
\quad\textit{$\Delta$ vs.\ FACE}
& \textit{+0} & \textit{$-$12} & \textit{+16} & \textit{+19} & \textit{$-$6} & \textit{$-$9}
& \textit{+9} & \textit{+25} & \textit{+3} & \textit{+9} & \textit{+31} & \textit{+0}
& \textit{+7} & \textit{0/12} \\
\quad\textit{$\Delta$ vs.\ BODY}
& \textit{+12} & \textit{+28} & \textit{+19} & \textit{+28} & \textit{+25} & \textit{+9}
& \textit{+41$^{*}$} & \textit{+28} & \textit{+22} & \textit{+34$^{*}$} & \textit{+28} & \textit{+16}
& \textit{+24} & \textit{2/12} \\
FACE\_TRANSITION & 72 & 81 & 69 & 66 & 66 & 75 & 56 & 56 & 69 & 69 & 31 & 34 & 62 & \\
\addlinespace[3pt]
\multicolumn{15}{@{}l}{\textit{Context with and without a character reaction}} \\
SCENE & 34 & 41 & 25 & 34 & 16 & 19 & 19 & 19 & 19 & 6 & 9 & 12 & 21 & \\
SCENE\_REACT & 66 & 66 & 75 & 66 & 75 & 66 & 69 & 78 & 62 & 59 & 59 & 47 & 66 & \\
\quad\textit{$\Delta$ vs.\ SCENE}
& \textit{+31} & \textit{+25} & \textit{+50$^{*}$} & \textit{+31} & \textit{+59$^{*}$} & \textit{+47$^{*}$}
& \textit{+50$^{*}$} & \textit{+59$^{*}$} & \textit{+44$^{*}$} & \textit{+53$^{*}$} & \textit{+50$^{*}$} & \textit{+34$^{*}$}
& \textit{+45} & \textit{9/12} \\
EVENT & 34 & 31 & 31 & 41 & 31 & 16 & 16 & 12 & 16 & 3 & 3 & 16 & 21 & \\
EVENT\_REACT & 66 & 69 & 84 & 66 & 72 & 62 & 81 & 72 & 62 & 72 & 59 & 38 & 67 & \\
\quad\textit{$\Delta$ vs.\ EVENT}
& \textit{+31} & \textit{+38} & \textit{+53$^{*}$} & \textit{+25} & \textit{+41} & \textit{+47$^{*}$}
& \textit{+66$^{*}$} & \textit{+59$^{*}$} & \textit{+47$^{*}$} & \textit{+69$^{*}$} & \textit{+56$^{*}$} & \textit{+22}
& \textit{+46} & \textit{7/12} \\
ALL\_DYNAMIC & 69 & 66 & 72 & 69 & 62 & 53 & 62 & 78 & 59 & 75 & 53 & 50 & 64 & \\
\addlinespace[3pt]
\multicolumn{15}{@{}l}{\textit{Atmosphere, character present: experiential vs.\ cinematic}} \\
CONTEXT\_MOOD & 47 & 41 & 31 & 41 & 28 & 22 & 22 & 28 & 22 & 6 & 25 & 9 & 27 & \\
CONTEXT\_HARD & 28 & 19 & 12 & 9 & 12 & 6 & 22 & 9 & 19 & 6 & 0 & 12 & 13 & \\
\quad\textit{$\Delta$ vs.\ CONTEXT\_MOOD}
& \textit{$-$19} & \textit{$-$22} & \textit{$-$19} & \textit{$-$31$^{*}$} & \textit{$-$16} & \textit{$-$16}
& \textit{+0} & \textit{$-$19} & \textit{$-$3} & \textit{+0} & \textit{$-$25$^{*}$} & \textit{+3}
& \textit{$-$14} & \textit{2/12} \\
\addlinespace[3pt]
\multicolumn{15}{@{}l}{\textit{Atmosphere, empty scene: experiential vs.\ cinematic}} \\
EMPTY\_MOOD & 34 & 47 & 44 & 47 & 44 & 41 & 38 & 41 & 31 & 25 & 25 & 38 & 38 & \\
EMPTY\_HARD & 50 & 50 & 41 & 38 & 41 & 28 & 38 & 41 & 38 & 25 & 25 & 31 & 37 & \\
\quad\textit{$\Delta$ vs.\ EMPTY\_MOOD}
& \textit{+16} & \textit{+3} & \textit{$-$3} & \textit{$-$9} & \textit{$-$3} & \textit{$-$12}
& \textit{+0} & \textit{+0} & \textit{+6} & \textit{+0} & \textit{+0} & \textit{$-$6}
& \textit{$-$1} & \textit{0/12} \\
\bottomrule
\end{tabular}
}
\vspace{2pt}

\parbox{\textwidth}{\scriptsize
Template definitions are given in Table~\ref{tab:appendix-templates}. Each
cell has $N=32$ and a confidence interval of roughly $\pm 17$ percentage points, so
nonsignificant differences are inconclusive rather than evidence of no effect.
$\Delta$ values are computed from the underlying counts and may differ by one
point from the difference of the rounded percentages shown. The two $\Delta$
rows under FACE\_BODY share the same target template and are not independent
tests. Target-emotion accuracy alone is
not reported here because it saturates: four templates reach a 12-model
mean of at least 85\%, leaving little room to separate them.}
\end{table*}

\end{document}